\documentclass[11pt]{article}

\usepackage[preprint]{acl}

\usepackage{booktabs}
\usepackage{tabularx}
\usepackage{times}
\usepackage{latexsym}
\usepackage{multirow}
\usepackage{multicol}
\usepackage[T1]{fontenc}
\usepackage[normalem]{ulem}

\usepackage[utf8]{inputenc}

\usepackage{microtype}
\usepackage[most]{tcolorbox}
\usepackage{inconsolata}

\usepackage{graphicx}
\usepackage[T1]{fontenc}
\usepackage{hyperref}
\usepackage{url}
\usepackage{booktabs}
\usepackage{amsfonts}
\usepackage{amsthm}
\usepackage{amsmath,amssymb,mathtools}
\usepackage{nicefrac}
\usepackage{microtype}
\usepackage{inconsolata}
\usepackage{latexsym}
\usepackage{graphicx}
\usepackage{subcaption}
\usepackage[most]{tcolorbox}
\usepackage{enumitem}
\usepackage{xcolor}
\usepackage{wrapfig}
\usepackage{amssymb}
\usepackage{array}
\usepackage[normalem]{ulem}
\usepackage{tabularx}
\usepackage{array}
\usepackage{enumitem}
\usepackage{longtable}
\usepackage{ragged2e}
\usepackage{placeins}

\newcolumntype{Y}{>{\RaggedRight\arraybackslash}X}

\usepackage{url}
\usepackage{tabularx}
\usepackage{makecell}

\setlist[itemize]{leftmargin=1.4em,itemsep=2pt,topsep=3pt,parsep=0pt}
\setlist[enumerate]{leftmargin=1.6em,itemsep=2pt,topsep=3pt,parsep=0pt}

\usepackage{multirow}

\usepackage{enumitem}
\setlist[itemize]{leftmargin=1.25em,itemsep=1pt,topsep=2pt,parsep=0pt}

\newcommand{\ours}{\textsc{WRIT}}

\usepackage{graphicx}

\title{%
  \hspace{-1em}%
  \smash{\raisebox{-1.45em}{\includegraphics[height=3.3em]{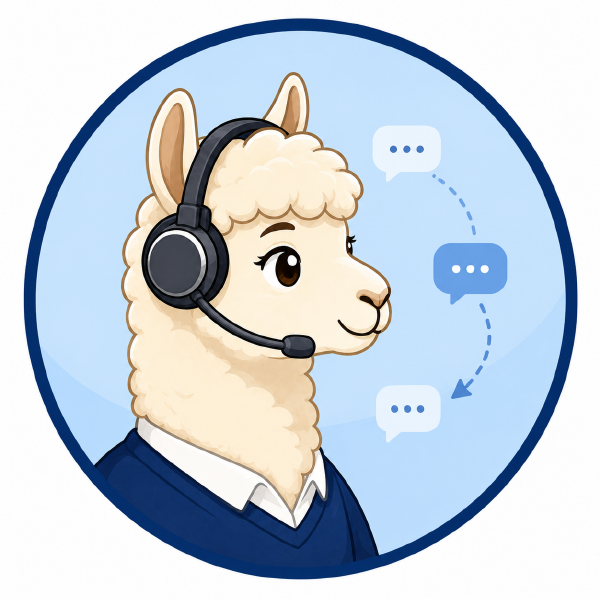}}}%
  \hspace{0.3em}%
  \ours{}: Write-Read Intensive Trajectory Synthesis\\ for Multi-Turn User-Facing Agents
}
\usepackage{fontawesome5}   % \faGlobe \faDatabase \faCube
\author{
  Hengrui Gu\textsuperscript{1} \quad
  Xiaotian Han\textsuperscript{2} \quad
  Kaixiong Zhou\textsuperscript{1} \\[0.5em]
  \textsuperscript{1}North Carolina State University \qquad
  \textsuperscript{2}Case Western Reserve University \\[0.3em]
  \texttt{hgu6@ncsu.edu} \quad \texttt{xhan@case.edu} \quad \texttt{kzhou22@ncsu.edu} \\[0.6em]
  \faGlobe\;Homepage: \href{https://hengrui-gu.github.io/WRIT/}{\texttt{https://hengrui-gu.github.io/WRIT/}} \\[0.3em]
  \faDatabase\;Dataset: \href{https://huggingface.co/datasets/Henryoung/WRIT-2K}{\texttt{https://huggingface.co/datasets/Henryoung/WRIT-2K}} \\[0.3em]
  \faCube\;Model: \href{https://huggingface.co/collections/Henryoung/writ-models}{\texttt{https://huggingface.co/collections/Henryoung/WRIT-models}}
}
\begin{document}
\maketitle
\begin{abstract}
Multi-turn user-facing agents must infer user intent from incomplete requests,
collect missing information through dialogue and tools, and execute valid actions. A training trajectory records this
process as an interleaved sequence of user messages, agent responses, tool
calls, etc. Synthesizing sufficiently complex trajectory has become a central
route to train agents: existing pipelines often increase difficulty by composing multiple user requests into longer tasks, producing
write-intensive trajectories that train sequential execution. 

We argue that a single write decision can
itself be difficult when the agent must gather and compare substantial read-tool
evidence before its arguments become identifiable, a challenge that
write-intensive data alone cannot address. Guided by this insight, we propose
\ours{} (\uline{W}rite-\uline{R}ead \uline{I}ntensive \uline{T}rajectory
Synthesis), a pipeline for synthesizing multi-turn agent training trajectories
along two complexity axes: the number of write decisions in a task and the
evidence burden of each individual decision. \ours{} first generates
write-intensive and read-heavy tasks. It then diversifies user behavior instructions to
reflect realistic conversational variation, and finally simulates agent-user
interactions in an executable environment to produce complete training
trajectories. The resulting data trains agents not only for longer task
execution, but also for robust, evidence-grounded decision making under high
information load. With only 2K synthesized trajectories, a 4B model trained on \ours{} outperforms GPT-5.1 no-think on $\tau^2$-bench and substantially reduces inference-time token usage, showing that compact SFT data can convert part of expensive test-time reasoning into efficient agent behavior.
\end{abstract}

\begin{table*}[t]
\centering
\small
\setlength{\tabcolsep}{4pt}
\renewcommand{\arraystretch}{1.15}
\begin{tabularx}{\textwidth}{
  >{\centering\arraybackslash}p{0.29\textwidth}
  >{\raggedright\arraybackslash}X
  l
}
\toprule
\textbf{Write Action} & \textbf{Reason} & \textbf{Tool Calls} \\
\midrule
\multirow[c]{2}{=}{%
\texttt{book\_reservation(}\\
\texttt{~~user\_id="emma\_johnson\_7098",}\\
\texttt{~~origin="EWR",}\\
\texttt{~~destination="IAH",}\\
\texttt{~~flight\_type="one\_way",}\\
\texttt{~~cabin="business",}\\
\texttt{~~flights=[\{date="2024-05-25",}\\
\texttt{~~~~flight\_number="HAT188"\}],}\\
\texttt{~~...)}%
}
&
\textbf{Simple task:} ``I need to book a one-way business class flight from Newark to Houston on May 25. Please book the direct flight that departs at 8:00 AM and arrives at 11:30 AM.''
&
\makecell[lt]{%
\textcolor[HTML]{1F77B4}{\texttt{1 x get\_user\_details}}\\
\textcolor[HTML]{1F77B4}{\texttt{1 x search\_direct\_flight}}\\
\textcolor[HTML]{B45F06}{\texttt{1 x book\_reservation}}} \\
\cmidrule(lr){2-3}
&
\textbf{Read-heavy task:} ``I need to book a one-way business class flight from the New York area to Houston. I'm flexible between May 25 and May 26, and I can depart from either Newark or LaGuardia. Please book the fastest overall flight.''
&
\makecell[lt]{%
\textcolor[HTML]{1F77B4}{\texttt{1 x get\_user\_details}}\\
\textcolor[HTML]{1F77B4}{\texttt{4 x search\_direct\_flight}}\\
\textcolor[HTML]{1F77B4}{\texttt{4 x search\_onestop\_flight}}\\
\textcolor[HTML]{B45F06}{\texttt{1 x book\_reservation}}} \\
\bottomrule
\end{tabularx}
\caption{A simple task and a read-heavy task can share the same gold write action, while differing in the amount of read evidence required to determine its arguments. The simple task uses 2 read-tool calls before the write action, whereas the read-heavy task uses 9 read-tool calls before executing the same booking action. Read tools are shown in blue and write tools in orange.}
\label{tab:simple-vs-readgraph}

\vspace{-12pt}
\end{table*}

\section{Introduction}
\label{sec:introduction}

% Language agents equipped with tools are a promising interface for automating real-world operational workflows, moving LLMs from passive text generation toward interactive task execution~\citep{yao2022react,schick2023toolformer,wang2025mcpbench}. In user-facing domains, such agents can assist customers with practical service requests, such as booking flights, changing reservations, or processing returns.~\citep{yao2024taubench,barres2025tau2bench,lu2024toolsandbox,qian2025userbench,cheng2025travelbench}. These multi-turn task-completion settings require agents to infer the user's underlying intent from incomplete or evolving requests, actively collect the missing information through dialogue and tools, and eventually execute the state-changing action needed to satisfy the user's goal, such as booking a flight or modifying a reservation. At the same time, the agent must ensure that its actions remain valid under domain policy, refusing requests that violate business rules such as ineligible cancellations or modifications~\citep{zhao2025muarl,rana2025agentchangebench,burdisso2025sdialog}. Training such agents therefore requires high-quality interaction trajectories that demonstrate not only which tool to call, but also how to converse, read evidence, follow policy, and act only when the request is sufficiently grounded~\citep{prabhakar2025apigenmt,li2025simia,chen2026cove,wang2026trajectory2task}.

Language agents equipped with tools are becoming a practical interface for automating user-facing workflows, from booking flights to changing reservations and processing returns~\citep{lu2024toolsandbox,drouin2024workarena,wang2025mcpbench,fang2025environment,barres2025tau2bench,qian2025userbench,cheng2025travelbench,qin2025compass}. In these multi-turn settings, an agent must infer an incomplete or evolving user intent, ask clarifying questions, read external records, follow domain policy, and execute valid state-changing actions~\citep{lu2024toolsandbox,zhao2025muarl,rana2025agentchangebench,burdisso2025sdialog,zhang2024agentsafetybench}. A training trajectory records this process as an interleaved sequence of user messages, agent responses, tool calls, and tool observations. High-quality trajectories are therefore the supervision that teaches an agent when to ask, when to read, which tool to call, what evidence to trust, and when it is safe to write~\citep{zeng2025toolacemt,xu2025toucan,gao2026selfevolving}.

Since collecting such trajectories from humans is expensive, synthetic trajectory generation has become a central route for training tool-using agents. Existing work follows several routes: executable simulation pipelines roll out interactions between user and agent models~\citep{prabhakar2025apigenmt,chen2026cove,wang2026trajectory2task}; LLM-driven pipelines synthesize trajectories or simulate environment feedback without a complete backend~\citep{li2025simia}; and environment-scaling approaches construct many tool-use environments from which trajectories can be collected~\citep{fang2025environment}. Together, these methods expand the quantity and diversity of training data and improve benchmark performance for multi-turn tool-use agents.

Most existing synthesis pipelines increase complexity by composing multiple user requests or state-changing actions into longer tasks. This trains agents for multi-step execution, sequential decision making, and long-horizon stability. Yet these pipelines mainly teach agents to do more, while overlooking difficulty that arises before any action is taken. In realistic service scenarios, the hard part is often gathering and comparing enough read-tool evidence to determine what arguments an action should carry. Users rarely provide all necessary identifiers; instead, they express preferences and descriptions, leaving the agent to search broadly before committing a state change. This motivate a new data synthesis question: \begin{tcolorbox}[
  enhanced,
  colback=gray!6,
  colframe=gray!40,
  boxrule=0.5pt,
  arc=3pt,
  shadow={2pt}{-2pt}{0pt}{gray!30},
  left=6pt, right=6pt, top=4pt, bottom=4pt
]
\textit{Beyond teaching agents to act for longer, can we synthesize
trajectories that teach them to read more carefully before they act?}
\end{tcolorbox}

Table~\ref{tab:simple-vs-readgraph} makes this distinction concrete. Both tasks share the same gold write action, \path{book_reservation(...)}, so from a write-action perspective they are identical. The difference is what the agent must do before writing. In the simple task, the user specifies the target flight by departure and arrival time, so one local search is enough. In the read-heavy task, the user asks for the fastest overall flight across multiple dates and departure airports, so the agent must search every airport-date combination, compare all returned candidates, and recover the correct \path{flight_number}; the read-tool count rises from 2 to 9. An agent trained only on shallow lookups may fail on such requests because it never learned to plan broad search, integrate evidence, and defer commitment until the arguments are grounded. Read-heavy trajectories are therefore a structurally distinct form of training complexity.

\begin{figure*}[t]
    \centering
    \includegraphics[width=0.95\textwidth]{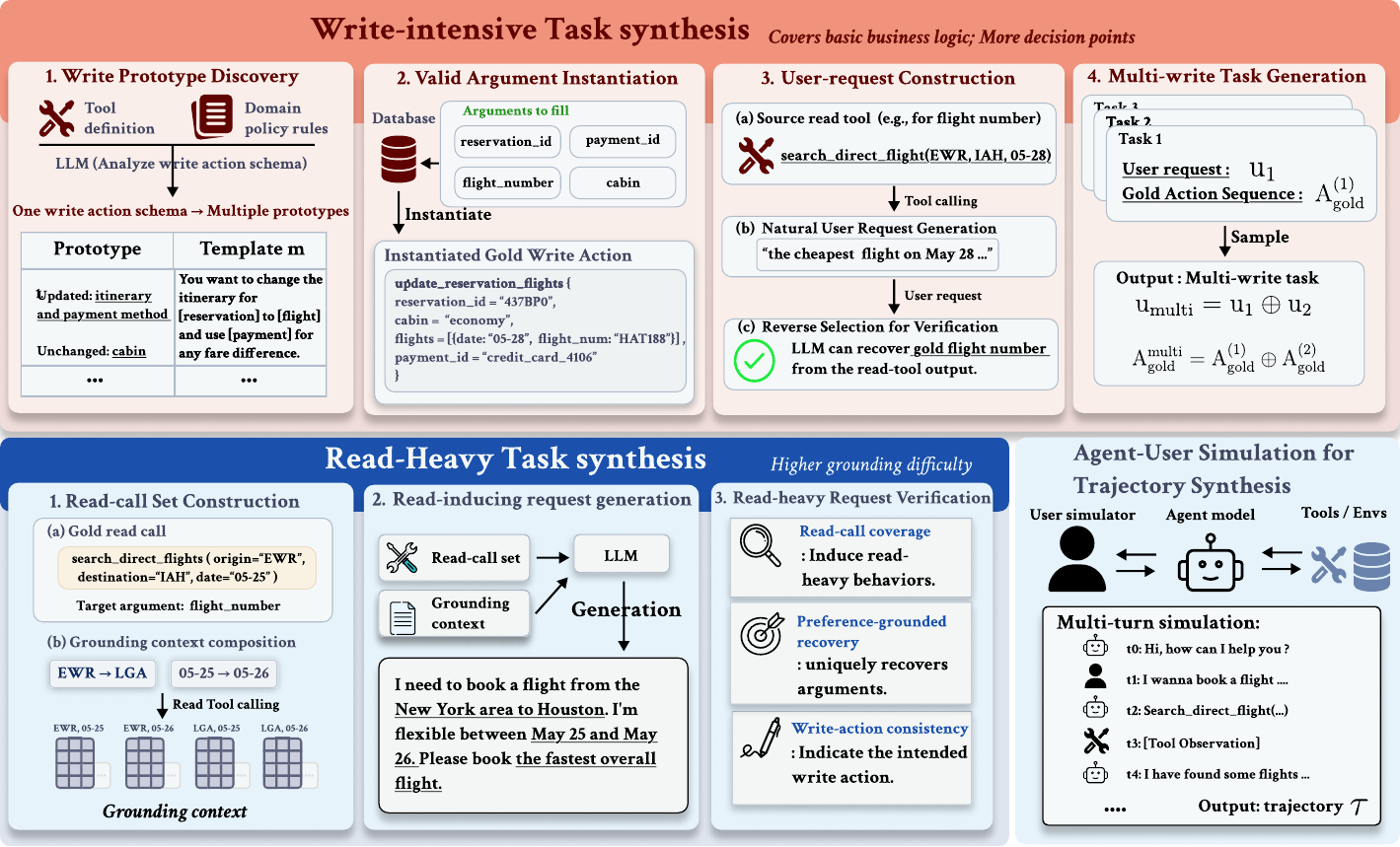}
    
    \vspace{-5pt}
    \caption{Overview of the \ours{} pipeline.}
    \label{fig:pipeline}
\vspace{-15pt}
\end{figure*}

Motivated by this observation, we propose \ours{} (\uline{W}rite-\uline{R}ead \uline{I}ntensive \uline{T}rajectory Synthesis), a pipeline that synthesizes training trajectories covering both action execution and evidence-intensive decision making. First, \ours{} generates service tasks with verifiable correct outcomes, spanning tasks with multiple sequential actions (i.e., write-intensive) and tasks where one action requires extensive reading and comparison (i.e., read-intensive). Second, \ours{} varies how users express and reveal the same request, so training data reflects realistic conversational behaviors rather than only cooperative, fully specified interactions. Third, \ours{} runs the agent and user through each task in an executable environment and retains successful interactions as complete training trajectories. Figure~\ref{fig:pipeline} summarizes this  pipeline.

We evaluate \ours{} on $\tau^2$-bench using a controlled 2K-trajectory training budget against strong synthetic-data baselines. 
\begin{itemize}[leftmargin=1.25em,itemsep=1pt,topsep=2pt,parsep=0pt]
\item \ours{} consistently outperforms prior trajectory synthesis methods across all three tested models (Qwen3-4B-Instruct-2507, Llama-3.1-8B-Instruct, Qwen2.5-14B-Instruct), with especially large gains on read-heavy task subsets. 
\item A 4B model trained with only 2K \ours{} trajectories outperforms GPT-5.1 no-think on $\tau^2$-bench and substantially narrows the gap to GPT-5.1 thinking, while using far fewer output tokens at inference time.
\item Ablations confirm that both read-heavy task synthesis and user-behavior diversification contribute independently. 
\end{itemize}
These results show that a small, carefully structured set of trajectories balancing write-intensive and read-intensive complexity can produce more capable and reliable agents than much larger but less structured datasets. Synthetic data should teach agents not only to act more, but also to know more before they act.

% [TBD] We evaluate \ours{} in the $\tau^2$-bench setting and on Tau-break, using a controlled 2K-sample training budget for fair comparison with strong synthetic-data baselines. Across pass@1 and pass@4 evaluation, \ours{} improves over prior trajectory synthesis methods, with ablations showing that both read-heavy grounding and script-guided interaction diversity contribute to agent performance. Further analyses indicate that read-heavy samples improve performance especially on tasks requiring substantial evidence gathering and comparison, supporting our central claim that synthetic agent data should control not only trajectory length, but also the information burden of individual decisions. 

\section{Problem Setup and Design Rationale}
\label{sec:prelim}

% We formalize our setting using $\tau^2$-bench as the running example, since it provides a canonical environment for multi-turn user-facing tasks with tools, policies, persistent database states, and executable success checks~\citep{barres2025tau2bench}. The same notation applies to other task-completion domains with structured tools and state-based evaluation.

% We study synthetic trajectory generation for \textit{multi-turn user-facing task-completion agents}~\citep{yao2024taubench,barres2025tau2bench}, where an agent completes a user's request through dialogue and tool use. These scenarios are challenging because user intents may be indirect, incomplete, revised over time, or infeasible under domain policy, while the agent must track dialogue state and ground the request into executable actions~\citep{yao2024taubench,barres2025tau2bench,prabhakar2025apigenmt,wang2026trajectory2task}. To make our method design concrete, we describe it in the environment setting of $\tau^2$-bench, since $\tau^2$-bench provides a canonical formulation of multi-turn user-facing agent tasks with tools, policies, persistent states, and executable success checks~\citep{barres2025tau2bench}.

% \subsection{Problem Setup}
% \label{subsec:problem_setup}

\noindent\textbf{2.1 Problem Setup.} We consider a user-facing operational domain, such as airline customer service, where an agent interacts with a user while operating over a database, a set of tools, and domain policy rules~\citep{yao2024taubench,barres2025tau2bench}. The tools include read tools, which observe the environment without changing it, such as \path{search_direct_flight(origin, destination, date)} for retrieving matching flight candidates. They also include write tools, which update the environment state, such as \path{book_reservation(user_id, origin, destination, flights, ...)} for creating a flight reservation. Domain policy rules constrain when write tools may be used, including rules such as \textit{``All reservations can be cancelled within 24 hours of booking.''}

A task specifies what the user wants the agent to accomplish and what a correct outcome looks like. We formalize a task as a tuple consisting of a user request $u$, an initial database state $s_{\mathrm{init}}$, a gold write-action sequence $A_{\mathrm{gold}}$, and a gold final database state $s_{\mathrm{gold}}$. Here, $u$ is the natural-language goal, $s_{\mathrm{init}}$ gives the starting conditions, $A_{\mathrm{gold}}$ specifies the correct state-changing actions, and $s_{\mathrm{gold}}$ is obtained by executing $A_{\mathrm{gold}}$ from $s_{\mathrm{init}}$ in a sandboxed environment. For a booking task, for example, $s_{\mathrm{gold}}$ is the database state after the correct reservation has been created, and task success is evaluated by checking whether the executed outcome matches $s_{\mathrm{gold}}$.

While the task defines what the agent must do, a training trajectory defines how the agent does it in a real conversation. A trajectory $\tau$ is the complete multi-turn interaction record generated by simulating the task, interleaving user messages, agent responses, tool calls, and tool observations across conversation turns. As supervised fine-tuning data, a trajectory teaches the agent when to ask for more information, which tool to call and with what arguments, how to interpret tool outputs, and when to execute a write action. Our pipeline first synthesizes tasks, then uses each task to simulate a trajectory, which lets us control task difficulty independently from how the trajectory unfolds.

\noindent\textbf{2.2 Two-Axis Trajectory Complexities.} \label{subsec:complexity} To synthesize useful training trajectories, we need to understand what makes a write decision difficult for the agent. The challenge is not only choosing the right write tool, but resolving the correct argument values from the user request, the conversation context, and tool observations; we call this process \textit{\textbf{argument grounding}}. For example, to book the right flight, the agent must determine the specific \path{flight_number} by reading flight search results, rather than being told it directly. Each write action is therefore a \textit{decision point}: before committing an action to environment, the agent must fully ground both the tool choice and its argument values.

This framing yields two independent ways to make agent training harder and more comprehensive. The first axis is the number of write decisions in a task: increasing it produces \textit{write-heavy trajectories} that train the agent on long-horizon sequential decision making. The second axis is the evidence burden of a single decision: increasing this axis produces \textit{read-heavy trajectories}, where one write action requires the agent to collect and compare multiple read-tool outputs before grounding its arguments. This second axis is important and comparatively underexplored: without read-heavy trajectories, an agent trained only on simple decisions may learn to act after a single lookup and fail when a real user's request requires searching across multiple options, dates, or alternatives before any valid write can be taken.

Our synthesis objective is therefore to generate training trajectories along both axes. % Read-heavy trajectories should present the agent with requests that require inspecting multiple tool outputs, where the correct write arguments are recoverable from evidence rather than directly stated in the user request. 
Together, write- and read-heavy trajectories teach the agent both long-horizon execution stability and evidence-intensive grounding under high information load. % In particular, read-heavy trajectories should induce a target set of read calls and remain solvable once the corresponding tool outputs are observed, so that the gold write arguments are recoverable from returned evidence rather than directly exposed in the user request.

\section{WRIT for Multi-turn Agent Training}
\label{sec:method}

Guided by this goal, we propose \ours{} 
(\uline{W}rite-\uline{R}ead \uline{I}ntensive \uline{T}rajectory Synthesis), 
a pipeline for generating multi-turn agent training data in three stages. First, \ours{} synthesizes write-read intensive tasks with known correct outcomes, covering both write-intensive service requests and read-heavy requests that require substantial evidence gathering. Second, \ours{} designs user behavior instructions that diversify how the user expresses and reveals the same underlying task across trajectories, so that training data reflects realistic conversational variation. Third, \ours{} runs the agent and user simulator through each task and behavior instruction in an executable environment, collecting successful interactions as complete supervised fine-tuning trajectories. In this workflow, the first two stages prepare the inputs, namely the task and behavior instruction, and the final stage turns them into training trajectories.

\subsection{Write-Read Intensive Task Synthesis}
\label{subsec:task_synthesis}

\ours{} first synthesizes tasks, each consisting of a user request $u$, an initial database state $s_{\mathrm{init}}$, a gold write-action sequence $A_{\mathrm{gold}}$, and a gold final state $s_{\mathrm{gold}}$. This subsection focuses entirely on task synthesis; the simulation that turns tasks into trajectories is introduced later in Section~\ref{subsec:rollout}. We control task complexity through following two branches.

\noindent\textbf{3.1.1 Write-intensive task synthesis.}
This branch synthesizes trajectories that cover the core write operations of the domain. Each trajectory trains the agent to identify common user intents, follow domain policy, and execute write actions with correctly grounded arguments. We describe the process in four steps.

\noindent\uline{\textit{Step 1: Write prototype discovery.}} The synthesis starts from identifying the popular write operations and user-facing scenarios the agent should learn to handle. We use an LLM to analyze the tool definitions and domain policy rules, and automatically derive a set of operation prototypes: each prototype captures a meaningful usage pattern for a write action and is paired with a natural-language template $m$ that describes the corresponding user intent with slots for grounded argument values. For example, one prototype for \path{update_reservation_flights(reservation_id, cabin, flights, payment_id)} captures the pattern where the user wants to change the itinerary and payment method, producing a template such as ``You want to change the itinerary for [reservation] to [flight] and use [payment] for any fare difference.'' These templates keep the generated user requests stable and semantically aligned with the target write action.

\noindent\uline{\textit{Step 2: Valid argument instantiation.}} This step
populates each prototype with concrete, valid argument values drawn from the current database state. For each prototype, we sample a feasible combination of database records that satisfies the prototype's constraints, such as selecting a user, one of the user's reservations, and a target cabin class that differs from the current one. This produces a fully instantiated gold write action $A_{\mathrm{gold}}$.

\noindent\uline{\textit{Step 3: User-request construction.}}
Based on the sampled write tool and arguments, we can construct a natural user request that expresses the intent behind the gold write action without directly exposing backend identifiers. Rather than inserting raw argument values, such as a flight number, into the request~\citep{prabhakar2025apigenmt}, we describe each argument through a natural preference, such as ``the cheapest flight'' instead of a literal flight ID~\citep{chen2026cove}, so the agent must read the environment to resolve it. % An LLM verifier then confirms that the description is unambiguous: given the relevant tool output, it must recover the original argument value. Only 
The verified descriptions are accepted and filled into the natural-language template $m$ to form the final user request $u$.

\noindent\uline{\textit{Step 4: Multi-write task generation.}}
We finally extend single-decision trajectories into multi-step trajectories that require the agent to complete several sequential write actions. For example, we combine two write-intensive trajectories by concatenating their user requests and gold write sequences into a single compound trajectory, i.e., $u_{\mathrm{multi}} = u_1 \oplus u_2$ and $A_{\mathrm{gold}}^{\mathrm{multi}} = A_{\mathrm{gold}}^{(1)} \oplus A_{\mathrm{gold}}^{(2)}$, with programmatic checks to prevent unintended execution conflicts. The resulting multi-write trajectories challenge the agent to sustain correct decision making across multiple decision points without losing track of the user's overall goal.

% \vspace{0.2em}
\noindent\textbf{3.1.2 Read-heavy task synthesis.} This branch synthesizes tasks in which a single write decision requires the agent to gather and compare evidence from multiple read-tool calls before the correct argument can be determined. Unlike write-intensive tasks, where arguments can be resolved through a small number of direct lookups, read-heavy tasks force the agent to search broadly, compare candidates across multiple tool outputs, and select the correct argument based on the user's preference. The construction proceeds in three steps.

\noindent\uline{\textit{Step 1: Read-call set construction.}} The synthesis process must first determines the full set of read-tool calls the agent should make to resolve the target write argument, and collect their outputs as an evidence pool. Starting from an instantiated gold write action, we identify one argument as the read-heavy target, i.e., the value the agent must discover through tool use. We identify the single read-tool call that contains this argument, which we call the gold read call, e.g., \path{search_direct_flight(origin="EWR", destination="IAH", date="2024-05-25")} in Table~\ref{tab:simple-vs-readgraph}, then generate perturbed variants of it by varying parameters such as date or departure airport. The outputs of all these calls form the \textit{grounding context}: the evidence pool the agent must compare to find the correct value.

\noindent\uline{\textit{Step 2:  Read-inducing request generation.}} It then generates a natural user request that requires the agent to consult the full evidence pool rather than stopping at a single lookup. An LLM generates the user request given the read-call set and grounding context, under two requirements: the stated user preference must lead the agent to consult all specified read-tool outputs, and it must uniquely identify the correct gold argument from the returned evidence. For example, ``fastest overall flight'' requires comparing candidates across all the searched airport-date combinations, as shown in Table~\ref{tab:simple-vs-readgraph}.

\noindent\uline{\textit{Step 3: Read-heavy request verification.}}
Finally, we verify that the generated request actually induces the intended evidence-gathering behavior and remains solvable. An LLM verifier checks following three properties: % \textit{read-call coverage}, where the request implies all read calls in the predefined set; \textit{preference-grounded recovery}, where the gold argument is recoverable from the grounding context  given the stated preference; and \textit{write-action consistency}, where the request clearly indicates the intended write action.
\begin{itemize}[leftmargin=1.25em,itemsep=1pt,topsep=2pt,parsep=0pt]
    \item \textit{Read-call coverage:} the request should imply all lookup operations in the read-call tool set.

    \item \textit{Preference-grounded recovery:} the verifier should recover gold argument from grounding context based on the stated user preference.

    \item \textit{Write-action consistency:} the request should clearly indicate the intended write action while leaving the read-heavy target argument to be resolved from evidence.
\end{itemize}
Requests that fail any check are discarded. The read-heavy task synthesis branch has now produced a verified user request $u$ and gold write-action sequence $A_{\mathrm{gold}}$.

\noindent\textbf{3.1.3 Gold-state construction.} After either synthesis branch produces a user request $u$ and a gold write-action sequence $A_{\mathrm{gold}}$, both branches enter the same gold-state construction step. We execute $A_{\mathrm{gold}}$ in a sandboxed environment initialized with $s_{\mathrm{init}}$ to obtain the gold final state $s_{\mathrm{gold}}$.The resulting state provides the executable supervision signal used later to verify whether a simulated trajectory actually completes the intended task.

\subsection{User Behavior Diversification}
\label{subsec:script}
Diversifying user behavior across trajectories is essential for training agents that remain robust when real users express the same request in different ways~\citep{ferreira2024mtad}.  The same task can unfold into many different conversations depending on how the user behaves: a user may reveal information gradually, correct a mistake mid-conversation, or add irrelevant small talk~\citep{hu2025impoliteusers}. If the training trajectories always assume a cooperative and information-complete user, the agent may be fragile at test time. User behavioral variation changes the conversational path but not the underlying goal or correct write action~\citep{hu2025impoliteusers}. This means we can explicitly diversify user behavior without changing the task's supervision signal: $A_{\mathrm{gold}}$ and $s_{\mathrm{gold}}$ stay fixed.

\ours{} maintains a library of reusable behavior instruction primitives, each describing a specific user behavior pattern. General task-completion primitives cover behaviors that arise in ordinary service conversations, including progressive disclosure, where the user reveals information gradually; self-correction, where the user fixes a stated value after a challenge; confirmation hesitation, where the user verifies the agent's summary before agreeing; mild emotion; and irrelevant asides~\citep{algherairy2025promptingusersim}. Policy-robustness primitives cover behaviors that specifically pressure the agent's policy boundary, including false-premise assertions, assume-style pressure, prior-agent approval claims, complaint pressure, and social flattery~\citep{hu2025impoliteusers}. These are needed because policy-sensitive tasks require the agent to refuse or redirect gracefully under adversarial user strategies that ordinary task-completion primitives do not cover.

For each synthesized task, we select a small number of compatible primitives from the library and prompts an LLM to instantiate them as concrete user-simulator instructions tailored to that task; for example, the instruction may ask the user simulator to initially give the wrong date and correct it only after the agent challenges it. These instructions govern only interaction style, namely how and when the user reveals information, not task content, namely what the user wants or which write action should be executed. The instructions are passed to the user simulator alongside the task request $u$ in Section~\ref{subsec:rollout}. Appendix~\ref{app:script-primitives} lists the script primitives used in our implementation and provides concrete examples of instantiated scripts.

\subsection{Trajectory Simulation and  Filtering}
\label{subsec:rollout}
This is where the synthesized task and user behavior instructions come together to produce complete training trajectories~\citep{prabhakar2025apigenmt,fang2025environment}. Section~\ref{subsec:task_synthesis} provides the task $(u, s_{\mathrm{init}}, A_{\mathrm{gold}}, s_{\mathrm{gold}})$, and Section~\ref{subsec:script} provides the user behavior instructions. With these inputs, we initialize the executable environment at $s_{\mathrm{init}}$ and run two models simultaneously: a user simulator guided by the task request $u$ and the behavior instructions, and an agent model given the domain policy and tool definitions. They interact turn by turn: the user expresses requests according to the behavior instructions, the agent responds and issues tool calls, and the environment executes those calls until the task is completed or refused. The output is a complete trajectory $\tau$ interleaving user messages, agent responses, tool calls, and tool observations.

Because the agent model may make errors during simulation, not all trajectories successfully realize the intended task, so we filter the data to keep only correct and complete demonstrations. % We execute the trajectory in the environment and compare the resulting database state $s_{\mathrm{end}}$ against the gold state $s_{\mathrm{gold}}$. Only trajectories where $s_{\mathrm{end}}=s_{\mathrm{gold}}$ are retained, meaning the agent executed the right write actions with the correct arguments; trajectories containing tool execution errors are additionally discarded. 
The retained trajectories form the training corpus $\mathcal{T}$ used for supervised fine-tuning. Since each retained trajectory comes from either a write-intensive or read-heavy task with a verified gold outcome, $\mathcal{T}$ systematically covers both axes of complexity defined in Section~\ref{subsec:complexity}.
Additional simulation details are provided in Appendix~\ref{app:simulation-details}.

\section{Experiments}
\label{sec:experiments}

\begin{table*}[t]
\centering
\small
\setlength{\tabcolsep}{2.4pt}
\renewcommand{\arraystretch}{1.08}
\begin{tabular}{llcccccccccc}
\toprule
\multirow{2}{*}{Model} & \multirow{2}{*}{Dataset}
& \multicolumn{2}{c}{$\tau^2$ Retail}
& \multicolumn{2}{c}{$\tau^2$ Airline}
& \multicolumn{2}{c}{$\tau^2$ Average}
& \multicolumn{2}{c}{$\tau^2$ Retail-Hard}
& \multicolumn{2}{c}{$\tau^2$ Airline-Hard} \\
\cmidrule(lr){3-4} \cmidrule(lr){5-6} \cmidrule(lr){7-8} \cmidrule(lr){9-10} \cmidrule(lr){11-12}
& & Pass$^1$ & Pass$^4$ & Pass$^1$ & Pass$^4$ & Pass$^1$ & Pass$^4$ & Pass$^1$ & Pass$^4$ & Pass$^1$ & Pass$^4$ \\
\midrule
\multirow{5}{*}{\shortstack{Qwen3-4B\\-Instruct-2507}}
& APIGen-MT
& 50.00{\scriptsize$\pm$2.77} & 23.68
& 20.00{\scriptsize$\pm$4.90} & 6.00
& 40.85{\scriptsize$\pm$2.28} & 18.29
& 43.15{\scriptsize$\pm$2.03} & 14.52
& 17.50{\scriptsize$\pm$6.45} & 5.00 \\
& Simia
& 53.73{\scriptsize$\pm$2.62} & 25.44
& 31.00{\scriptsize$\pm$6.63} & 10.00
& 46.80{\scriptsize$\pm$2.36} & 20.73
& 42.74{\scriptsize$\pm$4.06} & 14.52
& 21.25{\scriptsize$\pm$7.50} & 0.00 \\
& CoVe
& 59.65{\scriptsize$\pm$1.60} & 31.58
& 37.50{\scriptsize$\pm$6.40} & 20.00
& 52.90{\scriptsize$\pm$2.46} & 28.05
& 53.63{\scriptsize$\pm$2.42} & 22.58
& 33.75{\scriptsize$\pm$6.29} & 20.00 \\
& AReaL
& 59.43{\scriptsize$\pm$5.13} & 32.46
& 47.00{\scriptsize$\pm$3.46} & 36.00
& 55.64{\scriptsize$\pm$3.60} & 33.54
& 52.42{\scriptsize$\pm$11.82} & 25.81
& 42.50{\scriptsize$\pm$8.66} & 25.00 \\
& \ours{}
& \textbf{71.05}{\scriptsize$\pm$1.24} & \textbf{47.37}
& \textbf{61.00}{\scriptsize$\pm$3.83} & \textbf{42.00}
& \textbf{67.99}{\scriptsize$\pm$1.90} & \textbf{45.73}
& \textbf{66.13}{\scriptsize$\pm$2.28} & \textbf{38.71}
& \textbf{57.50}{\scriptsize$\pm$6.45} & \textbf{40.00} \\
\midrule
\multirow{5}{*}{\shortstack{Llama-3.1-8B\\-Instruct}}
& APIGen-MT
& 42.98{\scriptsize$\pm$1.75} & 18.42
& 20.50{\scriptsize$\pm$4.12} & 6.00
& 36.13{\scriptsize$\pm$2.19} & 14.63
& 34.27{\scriptsize$\pm$0.81} & 9.68
& 22.50{\scriptsize$\pm$5.00} & 5.00 \\
& Simia
& 40.79{\scriptsize$\pm$3.40} & 17.54
& 23.00{\scriptsize$\pm$2.00} & 8.00
& 35.37{\scriptsize$\pm$2.49} & 14.63
& 33.47{\scriptsize$\pm$3.58} & 12.90
& 16.25{\scriptsize$\pm$4.79} & 10.00 \\
& CoVe
& 52.19{\scriptsize$\pm$1.13} & 27.19
& 32.00{\scriptsize$\pm$4.90} & 14.00
& 46.04{\scriptsize$\pm$1.17} & 23.17
& \textbf{51.21}{\scriptsize$\pm$1.54} & 25.81
& 27.50{\scriptsize$\pm$2.89} & 15.00 \\
& AReaL
& 45.18{\scriptsize$\pm$0.51} & 20.18
& 43.50{\scriptsize$\pm$3.42} & 24.00
& 44.66{\scriptsize$\pm$0.77} & 21.34
& 37.50{\scriptsize$\pm$4.63} & 17.74
& 38.75{\scriptsize$\pm$2.50} & 15.00 \\
& \ours{}
& \textbf{54.61}{\scriptsize$\pm$2.90} & \textbf{31.58}
& \textbf{50.00}{\scriptsize$\pm$5.89} & \textbf{32.00}
& \textbf{53.20}{\scriptsize$\pm$3.32} & \textbf{31.71}
& 47.58{\scriptsize$\pm$5.01} & \textbf{27.42}
& \textbf{46.25}{\scriptsize$\pm$7.50} & \textbf{30.00} \\
\midrule
\multirow{5}{*}{\shortstack{Qwen2.5-14B\\-Instruct}}
& APIGen-MT
& 50.00{\scriptsize$\pm$3.72} & 24.56
& 27.00{\scriptsize$\pm$2.00} & 12.00
& 42.99{\scriptsize$\pm$2.84} & 20.73
& 43.15{\scriptsize$\pm$5.65} & 22.58
& 17.50{\scriptsize$\pm$5.00} & 5.00 \\
& Simia
& 51.10{\scriptsize$\pm$2.42} & 28.95
& 34.50{\scriptsize$\pm$1.91} & 18.00
& 46.04{\scriptsize$\pm$1.76} & 25.61
& 40.32{\scriptsize$\pm$3.95} & 17.74
& 23.75{\scriptsize$\pm$4.79} & 10.00 \\
& CoVe
& 58.11{\scriptsize$\pm$4.08} & 31.58
& 34.50{\scriptsize$\pm$3.00} & 16.00
& 50.91{\scriptsize$\pm$3.61} & 26.83
& 53.63{\scriptsize$\pm$5.49} & 27.42
& 30.00{\scriptsize$\pm$4.08} & 10.00 \\
& AReaL
& 57.68{\scriptsize$\pm$5.03} & 31.58
& 43.00{\scriptsize$\pm$2.58} & 28.00
& 53.20{\scriptsize$\pm$3.90} & 30.49
& 50.81{\scriptsize$\pm$6.52} & 29.03
& 30.00{\scriptsize$\pm$5.77} & 10.00 \\
& \ours{}
& \textbf{72.37}{\scriptsize$\pm$1.68} & \textbf{47.37}
& \textbf{57.50}{\scriptsize$\pm$4.43} & \textbf{38.00}
& \textbf{67.84}{\scriptsize$\pm$2.51} & \textbf{44.51}
& \textbf{66.13}{\scriptsize$\pm$2.28} & \textbf{37.10}
& \textbf{46.25}{\scriptsize$\pm$10.31} & \textbf{20.00} \\
\bottomrule
\end{tabular}
\caption{Tau2-bench evaluation results. $\tau^2$ Retail/Airline report success over all domain tasks, $\tau^2$ Average is task-count weighted across Retail and Airline, and Retail-Hard/Airline-Hard report fixed read-heavy subsets where the hardest decision point requires about six or more read/search calls. The exact hard-subset task groupings are listed in Appendix~\ref{app:tau2_read_heavy_subsets}. Pass$^1$ includes sample standard deviation across four trials. All numbers are percentages.}
\label{tab:main_tau2_hard_results}
\vspace{-12pt}
\end{table*}

\noindent\textbf{Experimental setting.}
We study in-domain specialization for agentic tasks, where training trajectories are synthesized using the predefined tool definitions and domain policies of an operational environment, and evaluation is conducted on unseen tasks within the same environment. This setup reflects deployment scenarios in which agents are specialized to known operational requirements before serving new user requests. We conduct our experiments under this setting, using the retail and airline domains of $\tau^2$-bench as the operational environments for synthesizing training trajectories. Both domains provide executable tools, domain policies, database states, and state-based success checks~\citep{barres2025tau2bench}.

\noindent\textbf{Training data.}
Using WRIT, we synthesize 2K training trajectories with balanced domain coverage, including 1K trajectories for the $\tau^2$-bench retail domain and 1K trajectories for the airline domain. Details of the trajectory synthesis setup are provided in Appendix~\ref{app:simulation-details}.

\noindent\textbf{Baselines.}
We compare WRIT with four synthetic trajectory generation methods for multi-turn user-facing agents: APIGen-MT~\citep{prabhakar2025apigenmt}, Simia~\citep{li2025simia}, CoVe~\citep{chen2026cove}, and AReaL~\citep{gao2026selfevolving}. For each baseline, we use its publicly released dataset synthesized for the $\tau^2$-bench retail and airline domains using the corresponding tool definitions and domain policy rules. These methods represent different synthesis strategies, including simulated agent--user interaction, seed-set expansion with simulated environment feedback, rule-based argument transformation, and LLM-controlled synthetic data generation. To compare data synthesis recipes under the same supervised fine-tuning budget, we use a controlled 2K trajectory-level setting for all main experiments. For public baselines with larger released datasets, we uniformly sample 2K trajectories at the trajectory level. This protocol isolates the effect of trajectory quality and task composition from the effect of dataset scale; we additionally report full-size baseline comparisons in Appendix~\ref{app:tau2_full_size_dataset_comparison}.

\noindent\textbf{Evaluation.}
We evaluate on $\tau^2$-bench, covering both retail and airline domains. In addition to the full task sets, we report performance on fixed read-heavy subsets where the hardest decision point requires about six or more read/search calls; the subset definitions are provided in Appendix~\ref{app:tau2_read_heavy_subsets}. We use the $\mathrm{Pass}^k$ reliability metric~\citep{yao2024taubench}. For each task $i$, we run $n$ independent trials and let $c_i$ denote the number of successful trials. The $\mathrm{Pass}^k$ score is computed as $\frac{1}{|\mathcal{Q}|}\sum_{i\in\mathcal{Q}}\binom{c_i}{k}/\binom{n}{k}$, where $\mathcal{Q}$ is the evaluated task set. Intuitively, $\mathrm{Pass}^1$ is the average success rate over repeated trials, while $\mathrm{Pass}^k$ estimates the probability that $k$ randomly sampled trials for the same task all succeed. Larger $k$ therefore gives a stricter measure of reliability, because a model must solve the same task consistently rather than succeed only occasionally. In our experiments, we run each task four times and report $\mathrm{Pass}^1$ and $\mathrm{Pass}^4$. Higher $\mathrm{Pass}^k$ indicates more consistent behavior across repeated attempts.

\noindent\textbf{Models and implementation details.}
We focus on non-thinking agent settings, where the deployed model must act directly without explicit long-form reasoning. We therefore fine-tune multiple instruction-tuned base models, including Qwen3-4B-Instruct-2507, Llama-3.1-8B-Instruct, and Qwen2.5-14B-Instruct. For each base model and dataset, we perform full-parameter supervised fine-tuning on the corresponding training trajectories. Dataset statistics, training hyperparameters, and additional implementation details are provided in Appendix~\ref{app:experiment-details}.

\begin{table*}[t]
\centering
\small
\setlength{\tabcolsep}{2.8pt}
\renewcommand{\arraystretch}{1.08}
\begin{tabular}{lcccccccccc}
\toprule
\multirow{2}{*}{Variant}
& \multicolumn{2}{c}{$\tau^2$ Retail}
& \multicolumn{2}{c}{$\tau^2$ Airline}
& \multicolumn{2}{c}{$\tau^2$ Average}
& \multicolumn{2}{c}{$\tau^2$ Retail-Hard}
& \multicolumn{2}{c}{$\tau^2$ Airline-Hard} \\
\cmidrule(lr){2-3} \cmidrule(lr){4-5} \cmidrule(lr){6-7} \cmidrule(lr){8-9} \cmidrule(lr){10-11}
& Pass$^1$ & Pass$^4$ & Pass$^1$ & Pass$^4$ & Pass$^1$ & Pass$^4$ & Pass$^1$ & Pass$^4$ & Pass$^1$ & Pass$^4$ \\
\midrule
\ours
& \textbf{71.05}{\scriptsize$\pm$1.24} & \textbf{47.37}
& \textbf{61.00}{\scriptsize$\pm$3.83} & \textbf{42.00}
& \textbf{67.99}{\scriptsize$\pm$1.90} & \textbf{45.73}
& \textbf{66.13}{\scriptsize$\pm$2.28} & \textbf{38.71}
& \textbf{57.50}{\scriptsize$\pm$6.45} & \textbf{40.00} \\
\quad w/o read-heavy
& 67.11{\scriptsize$\pm$3.24} & 46.49
& 52.50{\scriptsize$\pm$5.74} & 30.00
& 62.65{\scriptsize$\pm$3.39} & 41.46
& 57.66{\scriptsize$\pm$3.05} & 33.87
& 41.25{\scriptsize$\pm$6.29} & 15.00 \\
\quad w/o script
& 69.96{\scriptsize$\pm$4.72} & 45.61
& 49.50{\scriptsize$\pm$6.61} & 26.00
& 63.72{\scriptsize$\pm$2.84} & 39.63
& 62.90{\scriptsize$\pm$8.22} & 32.26
& 45.00{\scriptsize$\pm$17.80} & 20.00 \\
\quad w/o multi-write
& 67.32{\scriptsize$\pm$3.31} & 43.86
& 55.00{\scriptsize$\pm$4.16} & 32.00
& 63.57{\scriptsize$\pm$1.15} & 40.24
& 59.27{\scriptsize$\pm$4.03} & 29.03
& 51.25{\scriptsize$\pm$8.54} & 30.00 \\
\bottomrule
\end{tabular}
\caption{Ablation results on $\tau^2$-bench using Qwen3-4B-Instruct-2507. We report Pass$^1$ and Pass$^4$ on the full Retail/Airline task sets, their task-count weighted average, and the fixed read-heavy hard subsets.}
\label{tab:tau2_ablation_results}

\vspace{-14pt}
\end{table*}

\subsection{Results and Analysis}
\label{subsec:results}

\noindent\textbf{\ours{} expands the agent capability boundary and improves reliability.}
Table~\ref{tab:main_tau2_hard_results} shows that \ours{} substantially improves multi-turn agent performance across model families. On Qwen3-4B-Instruct-2507, \ours{} achieves a $\tau^2$ Average Pass1 of 67.99, outperforming AReaL by 12.35 points, and improves Pass4 from 33.54 to 45.73. The same pattern holds for Llama-3.1-8B-Instruct, where \ours{} improves the average Pass1 from 46.04 with CoVe to 53.20, and for Qwen2.5-14B-Instruct, where \ours{} improves the average Pass1 from 53.20 with AReaL to 67.84. Higher Pass1 suggests that the trained agent can solve a broader set of tasks in a single attempt, while higher Pass4 indicates more stable behavior across repeated trials. Together, these gains show that \ours{} improves both capability coverage and reliability in multi-turn user-facing settings.

\begin{figure}[t]
\centering
\includegraphics[width=0.45\textwidth]{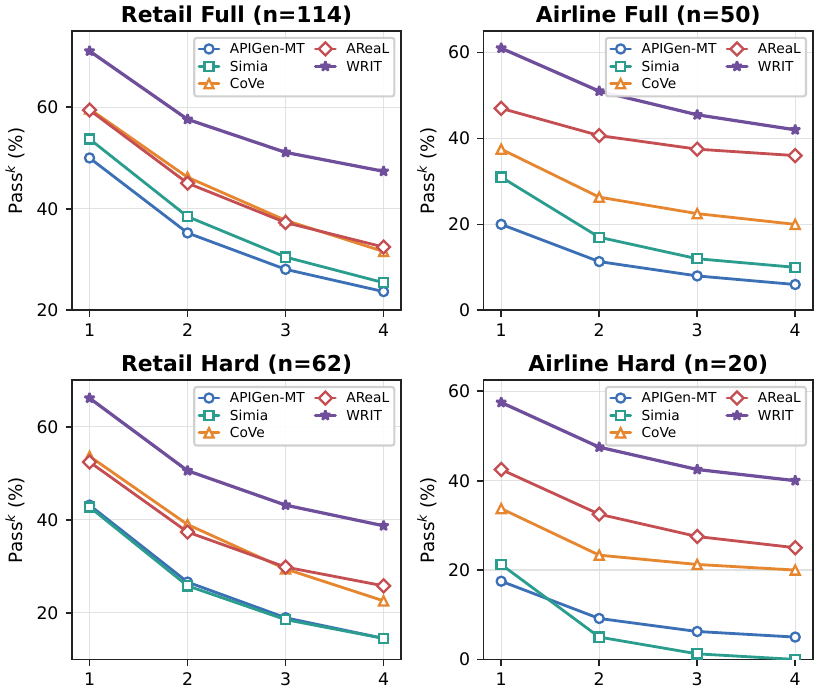}
\caption{Pass$^k$ curves for Qwen3-4B-Instruct-2507. The horizontal axis indexes $k=1,2,3,4$.}
\label{fig:qwen3_4b_tau2_passk_curves}
\vspace{-14pt}
\end{figure}

\noindent\textbf{Read-heavy synthesis addresses a key weakness of user-facing agents.}
The gains are especially clear on the read-heavy subsets, which correspond to difficult $\tau^2$-bench tasks requiring substantial read/search behavior before the final decision. For Qwen3-4B-Instruct-2507, \ours{} improves Airline-Hard Pass1 from 42.50 with AReaL to 57.50, and improves Pass4 from 25.00 to 40.00. On Retail-Hard, \ours{} also improves Pass1 from 53.63 with CoVe to 66.13. Similar improvements appear for other two base models, as also shown by the Pass$^k$ curves in Figure~\ref{fig:qwen3_4b_tau2_passk_curves}. These results suggest that our synthesized trajectories directly improve a capability gap in current user-facing agents: they need practice not only executing tools, but also gathering and comparing enough evidence before committing to a write action.

\begin{table}[t]
\centering
\small
\setlength{\tabcolsep}{3.0pt}
\renewcommand{\arraystretch}{1.08}
\resizebox{\columnwidth}{!}{
\begin{tabular}{lcccc}
\toprule
Method & Retail & Airline & Avg. & Output Tokens (USD) \\
\midrule
GPT-5.1 thinking & \textbf{82.46} & \textbf{72.00} & \textbf{79.27} & 1,520,619 (\$17.52) \\
GPT-5.1 no-think & 69.30 & 48.00 & 62.80 & 318,180 (\$5.56) \\
WRIT-4B           & \underline{71.05} & \underline{61.00} & \underline{67.99} & \textbf{251,405 (--)} \\
\bottomrule
\end{tabular}
}
\caption{GPT-5.1 evaluation results on $\tau^2$-bench. Retail, Airline, and Avg. are percentages. Output Tokens counts agent-side completion tokens for one full $\tau^2$ evaluation, with agent-side API cost shown in parentheses.}
\label{tab:gpt51_writ_pass1_tokens_cost}

\vspace{-15pt}
\end{table}

\begin{table*}[t]
\centering
\scriptsize
\renewcommand{\arraystretch}{1.12}
\resizebox{\textwidth}{!}{%
\begin{tabular}{lcccccccccc}
\toprule
& \multicolumn{5}{c}{Retail ($n=114$)}
& \multicolumn{5}{c}{Airline ($n=50$)} \\
\cmidrule(lr){2-6}
\cmidrule(lr){7-11}
Dataset
& \shortstack{$\geq 4$\\$(n=102)$}
& \shortstack{$\geq 5$\\$(n=84)$}
& \shortstack{$\geq 6$\\$(n=62)$}
& \shortstack{$\geq 7$\\$(n=29)$}
& \shortstack{$\geq 8$\\$(n=17)$}
& \shortstack{$\geq 4$\\$(n=28)$}
& \shortstack{$\geq 5$\\$(n=25)$}
& \shortstack{$\geq 6$\\$(n=20)$}
& \shortstack{$\geq 7$\\$(n=15)$}
& \shortstack{$\geq 8$\\$(n=7)$} \\
\midrule
APIGen-MT
& 48.53/21.57
& 45.24/16.67
& 43.15/14.52
& 37.07/6.90
& 22.06/0.00
& 16.07/3.57
& 15.00/4.00
& 17.50/5.00
& 13.33/6.67
& 21.43/14.29 \\
Simia
& 50.49/23.53
& 45.24/16.67
& 42.74/14.52
& 27.59/6.90
& 19.12/0.00
& 23.21/3.57
& 21.00/0.00
& 21.25/0.00
& 20.00/0.00
& 10.71/0.00 \\
CoVe
& 58.09/30.39
& 57.14/26.19
& 53.63/22.58
& 45.69/17.24
& 38.24/11.76
& 35.71/17.86
& 35.00/20.00
& 33.75/20.00
& 31.67/13.33
& 21.43/14.29 \\
AReaL
& 57.84/30.39
& 55.65/27.38
& 52.42/25.81
& 39.66/13.79
& 42.65/17.65
& 41.07/28.57
& 38.00/24.00
& 42.50/25.00
& 38.33/20.00
& 25.00/14.29 \\
\ours{}
& \textbf{70.10/45.10}
& \textbf{69.05/42.86}
& \textbf{66.13/38.71}
& \textbf{55.17/24.14}
& \textbf{45.59/23.53}
& \textbf{62.50/42.86}
& \textbf{59.00/40.00}
& \textbf{57.50/40.00}
& \textbf{55.00/40.00}
& \textbf{35.71/28.57} \\
\bottomrule
\end{tabular}%
}
\caption{Sensitivity to the read-heavy cutoff. Each cell reports
$\mathrm{Pass}^1/\mathrm{Pass}^4$ (\%) over four trials using
Qwen3-4B-Instruct-2507; $n$ denotes the number of tasks retained at
each cutoff. The default cutoff is six reads. }
\label{tab:read_cutoff_sensitivity}
\end{table*}

\noindent\textbf{Pipeline components specialize into complementary capabilities.}
Table~\ref{tab:tau2_ablation_results} shows that all three components contribute to the final performance. \textbf{Read-heavy grounding strengthens evidence-intensive decisions.} Removing read-heavy trajectories only mildly reduces Retail Pass1 from 71.05 to 67.11, but the drop becomes much larger on Retail-Hard, from 66.13 to 57.66. The effect is even more pronounced on Airline-Hard, where Pass1 drops from 57.50 to 41.25 and Pass4 collapses from 40.00 to 15.00. This pattern directly supports our main hypothesis: read-heavy samples do not merely improve general performance, but specifically improve the agent's capability and stability on difficult tasks that require substantial evidence gathering before acting. \textbf{Scripts improve robustness near the policy boundary.} Removing scripts causes the largest full-domain drop on Airline, reducing Pass1 from 61.00 to 49.50 and Pass4 from 42.00 to 26.00; it also substantially hurts Airline-Hard, where Pass4 falls from 40.00 to 20.00. The retail domain is less affected, suggesting that the script layer is especially valuable in policy-sensitive settings where adversarial user patterns, such as false premises, pressure, or delayed policy-relevant information, stress the agent's refusal and policy-following behavior. \textbf{Multi-write composition provides useful task-composition coverage.} Removing multi-write trajectories lowers the average Pass1 from 67.99 to 63.57 and Pass4 from 45.73 to 40.24, with the largest drop appearing on Retail-Hard Pass4, from 38.71 to 29.03. This indicates that compound tasks still provide important training signal for maintaining correctness across multiple requested operations. Overall, the ablations show that both complexity-oriented sample types, multi-write and read-heavy, substantially improve performance, while scripts mainly improve stability under user-side variation and policy-boundary stress.       

\noindent\textbf{\ours{} approaches strong API agents with substantially lower inference cost.}
Table~\ref{tab:gpt51_writ_pass1_tokens_cost} compares \ours{} with GPT-5.1 variants on $\tau^2$-bench. Although GPT-5.1 thinking achieves the highest score, it uses over 1.5M output tokens for one full Retail+Airline evaluation, reflecting the high inference cost of relying on test-time reasoning. In contrast, \ours{} outperforms GPT-5.1 no-think on both domains, improving the average Pass1 from 62.80 to 67.99, while using fewer output tokens. This suggests that our synthesized trajectories transfer part of the required evidence-gathering and policy-following behavior into the model parameters through SFT, allowing a smaller non-thinking agent to act more efficiently at inference time. The gap to GPT-5.1 thinking further indicates that explicit reasoning remains powerful, but \ours{} provides a cost-effective alternative when deployment requires direct, low-token agent behavior.

\begin{table}[t]
\centering
\small
\setlength{\tabcolsep}{3.5pt}
\begin{tabular}{lrrrrr}
\toprule
& \multicolumn{4}{c}{Volume per trajectory}
& \multicolumn{1}{c}{Evidence} \\
\cmidrule(lr){2-5}
\cmidrule(l){6-6}
Dataset & Turns & Calls & Writes & Tokens (K) & R/W \\
\midrule
APIGen-MT
& 18.46 & 4.40 & 1.09 & 7.52 & 2.78 \\
Simia
& 17.53 & 4.17 & 1.16 & 6.85 & 2.47 \\
CoVe
& 24.36 & 6.78 & 1.73 & 10.30 & 2.89 \\
AReaL
& \textbf{28.27} & \textbf{8.99} & \textbf{1.83}
& \textbf{12.64} & 3.87 \\
\ours{}
& 24.63 & 6.61 & 1.12 & 9.03 & \textbf{4.78} \\
\bottomrule
\end{tabular}
\caption{Full-dataset trajectory statistics (per-trajectory averages). Turns include user, assistant, and tool messages; Calls and Writes denote all tool calls and database-changing calls, respectively. Tokens are reported in thousands using the Qwen3-4B-Instruct-2507 tokenizer, and R/W denotes the ratio of mean read calls to mean write calls. Column maxima are bolded.}
\label{tab:dataset_statistics}
\end{table}

\noindent\textbf{Performance gains are robust to the read-heavy cutoff.}
Table~\ref{tab:read_cutoff_sensitivity} reports results across read-heavy cutoffs from four to eight reads. \ours{} achieves the highest $\mathrm{Pass}^1$ and $\mathrm{Pass}^4$ in both domains at every cutoff, showing that its advantage is not specific to the default subset definition. As the cutoff increases, the retained tasks become more read-intensive and the subset sizes decrease, but the relative performance pattern remains stable. We use six reads as the default because it corresponds to the median split of the benchmark, retaining 82 of 164 tasks, including 62 Retail and 20 Airline tasks. Raising the cutoff to seven reduces the evaluation set to 44 tasks, and raising it to eight leaves only 24 tasks, whereas lower cutoffs include a larger proportion of less read-intensive tasks. The six-read cutoff therefore provides a balance between read intensity and evaluation stability.

\noindent\textbf{\ours{} shifts trajectory complexity toward evidence gathering.}
Table~\ref{tab:dataset_statistics} compares the structural profiles of the synthesized datasets. AReaL is the largest on all volume-oriented measures, with the most turns, tool calls, write actions, and tokens per trajectory. CoVe is close to \ours{} in overall interaction volume, but contains substantially more write actions. In contrast, \ours{} is not the largest on any volume measure, yet has the highest evidence burden: 4.78 read calls per write call, compared with 3.87 for the next highest dataset. This profile shows that \ours{} shifts trajectory complexity toward gathering evidence before state-changing actions rather than simply increasing interaction length or operation count. By introducing read-heavy complexity as a complementary axis to write-oriented task composition, \ours{} provides compact supervision with more balanced coverage of evidence-gathering and state-changing capabilities. Together with the preceding ablations, this structural difference is consistent with the larger gains observed on read-heavy tasks.

\section{Conclusion}
\label{sec:conclusion}

We presented \ours{}, a trajectory synthesis pipeline for multi-turn user-facing agents that controls task complexity along two axes: the number of write decisions and the read evidence required to resolve each decision. By combining decision-coverage tasks, read-heavy grounding tasks, and scripted user behaviors, \ours{} produces clean SFT trajectories in executable environments. Experiments on $\tau^2$-bench show significant gains across models, especially on read-heavy hard subsets, demonstrating the importance of two-axis complexity control and opening a promising direction for future agentic trajectory synthesis.

\section*{Limitations}

\label{sec:limitations}

\paragraph{Compositional hard samples.}
\ours{} controls task complexity along two axes: increasing the number of decision points through multi-write tasks, and increasing the grounding difficulty of individual decision points through read-heavy tasks. In this work, we study these two axes mostly as separate sources of difficulty. We have not fully explored their composition, such as constructing multi-write tasks where each decision point is also read-heavy. Such samples may further stress long-horizon state tracking and evidence-intensive grounding at the same time.

\paragraph{Mixture of complexity types.}
Our training data contains both decision-coverage samples and read-heavy grounding samples, but we do not exhaustively study the optimal mixture ratio between different complexity types. Different model families or base capabilities may benefit from different proportions of low-read, multi-write, read-heavy, and policy-robust samples. A more systematic mixture study could clarify how each type of synthetic trajectory shapes agent behavior during supervised fine-tuning.

\section*{Acknowledgments}

We thank the anonymous reviewers for their constructive feedback, which helped improve this paper. This work was supported in part by the National Science Foundation (NSF) under Grants CNS-2431516 and IIS-2551751, and by the National Institutes of Health (NIH) under Grant 1OT2OD038051-01.

\bibliography{custom}

\clearpage
\appendix

\section{Related Work}
\label{app:related-work}

\subsection{Synthetic trajectories for agent training}

Recent agent research increasingly uses full interaction trajectories as supervision for teaching models agentic capabilities, rather than relying only on final-answer labels. In web and workflow environments, agent traces describe how models navigate interfaces, use tools, and complete multi-step tasks~\citep{zhou2023webarena,koh2024visualwebarena,drouin2024workarena,trivedi2024appworld}. In software engineering, trajectories capture repository navigation, code editing, tool execution, and issue resolution~\citep{jimenez2024swebench,yang2024sweagent}. Other work studies tool-use or function-calling traces across heterogeneous APIs and environments~\citep{patil2023gorilla,basu2024nestful,xu2025toucan,zeng2025toolacemt}. These studies show that trajectory-level supervision can teach models intermediate actions and tool interactions that are difficult to learn from final-answer labels alone.

\subsection{Trajectory synthesis for multi-turn user-facing agents}

Training multi-turn user-facing agents requires trajectories that capture dialogue state tracking, user intent clarification, policy adherence, tool use, and state-changing execution~\citep{yao2024taubench,barres2025tau2bench}. Recent work has therefore studied how to synthesize such trajectories without relying on large-scale human collection.

APIGen-MT proposes a two-phase pipeline that first constructs task blueprints with ground-truth actions and then realizes them as multi-turn interactions through simulated human-agent interplay~\citep{prabhakar2025apigenmt}. Simia expands small seed datasets into more diverse training trajectories, using reasoning models to simulate environment feedback and support data augmentation without a fully implemented executable backend~\citep{li2025simia}. CoVe focuses on rule-based argument transformation: it replaces directly exposed tool arguments with predefined indirect descriptions, so that the agent must recover the hidden argument through tool use~\citep{chen2026cove}. AReaL follows an LLM-controlled generation pipeline, where an LLM drives the construction of synthetic tasks, dialogues, tool calls, and executable checkers for multi-turn tool-use training~\citep{gao2026selfevolving}. AgentScaler broadens the setting by constructing many synthetic function-calling environments from which agent trajectories can be collected~\citep{fang2025environment}.

These works commonly increase task difficulty by composing multiple user requests or write actions into compound tasks. This produces longer trajectories and trains agents for long-horizon execution, but it mainly increases the number of decision points. Our work studies a complementary axis of complexity. Instead of only asking the agent to execute more write actions, \ours{} constructs read-heavy tasks where a single write action requires substantial read-tool evidence before its arguments can be resolved. This differs from fixed rule-based argument rewriting: \ours{} generates natural user requests that induce specified read-heavy behavior, and uses reverse-selection verification to ensure that the intended write argument remains recoverable from the returned evidence.

% Plain appendix snippet. No longtable or extra package is required.
% If the main paper has already entered appendix mode, remove \appendix.
% Plain appendix snippet using regular tables only.
% No longtable package is required.
% If the main paper has already entered appendix mode, remove \appendix.
% If the paper is single-column, table* can be changed to table.

\section{Full-Size Dataset Comparison on \texorpdfstring{$\tau^2$}{tau2}-Bench}
\label{app:tau2_full_size_dataset_comparison}

Our main experiments use uniformly sampled 2K trajectory-level training sets for all datasets. This design is intended to isolate data quality and task composition while holding the SFT data budget fixed across methods. However, several public baselines are released at larger scales, such as APIGen-MT-5K, Simia-90K, and CoVe-12K. A natural concern is therefore that the 2K sampling protocol could understate the performance of these baselines by discarding useful examples.

\begin{figure*}[t]
\centering
\includegraphics[width=\textwidth]{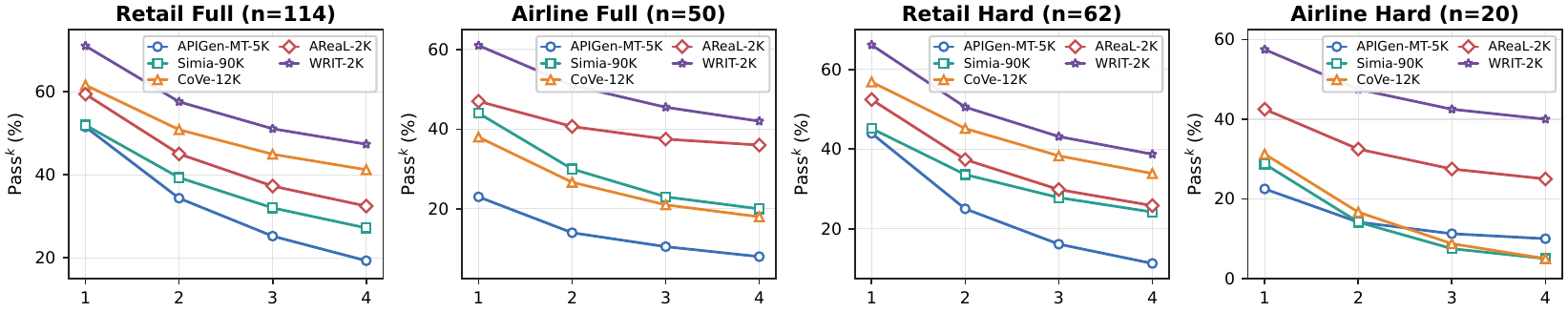}
\caption{Pass$^k$ degradation curves for the full-size dataset comparison on $\tau^2$-bench using Qwen3-4B-Instruct-2507. The horizontal axis indexes $k=1,2,3,4$; panel titles report the number of evaluated tasks. Unlike the controlled 2K-budget main comparison, this setting trains each dataset at its available full scale, including APIGen-MT-5K, Simia-90K, CoVe-12K, AReaL-2K, and \ours-2K.}
\label{fig:tau2_full_size_dataset_comparison_passk_curves}
\vspace{-5pt}
\end{figure*}

To address this possible confound, we run an additional full-size comparison using the same base model, training recipe, and evaluation protocol as the main Qwen3-4B-Instruct-2507 experiments. Specifically, we train on each dataset at its available full scale and evaluate with the same strict Pass$^k$ computation, where context-window and model-side failures are counted as incorrect. This experiment is not meant to replace the controlled 2K-budget comparison; rather, it verifies that our conclusions are not an artifact of uniformly downsampling larger baseline datasets. The full-size results are reported in Table~\ref{tab:tau2_full_size_dataset_comparison}, with the corresponding Pass$^k$ curves shown in Figure~\ref{fig:tau2_full_size_dataset_comparison_passk_curves}.

\begin{table*}[t]
\centering
\small
\setlength{\tabcolsep}{2.8pt}
\renewcommand{\arraystretch}{1.08}
\begin{tabular}{lcccccccccc}
\toprule
\multirow{2}{*}{Dataset}
& \multicolumn{2}{c}{$\tau^2$ Retail}
& \multicolumn{2}{c}{$\tau^2$ Airline}
& \multicolumn{2}{c}{$\tau^2$ Average}
& \multicolumn{2}{c}{$\tau^2$ Retail-Hard}
& \multicolumn{2}{c}{$\tau^2$ Airline-Hard} \\
\cmidrule(lr){2-3} \cmidrule(lr){4-5} \cmidrule(lr){6-7} \cmidrule(lr){8-9} \cmidrule(lr){10-11}
& Pass$^1$ & Pass$^4$ & Pass$^1$ & Pass$^4$ & Pass$^1$ & Pass$^4$ & Pass$^1$ & Pass$^4$ & Pass$^1$ & Pass$^4$ \\
\midrule
APIGen-MT-5K
& 51.54{\scriptsize$\pm$1.10} & 19.30
& 23.00{\scriptsize$\pm$3.46} & 8.00
& 42.84{\scriptsize$\pm$0.30} & 15.85
& 43.95{\scriptsize$\pm$2.42} & 11.29
& 22.50{\scriptsize$\pm$5.00} & 10.00 \\
Simia-90K
& 51.97{\scriptsize$\pm$3.54} & 27.19
& 44.00{\scriptsize$\pm$6.73} & 20.00
& 49.54{\scriptsize$\pm$2.74} & 25.00
& 45.16{\scriptsize$\pm$5.43} & 24.19
& 28.75{\scriptsize$\pm$11.81} & 5.00 \\
CoVe-12K
& 61.62{\scriptsize$\pm$4.01} & 41.23
& 38.00{\scriptsize$\pm$9.38} & 18.00
& 54.42{\scriptsize$\pm$4.06} & 34.15
& 56.85{\scriptsize$\pm$5.16} & 33.87
& 31.25{\scriptsize$\pm$14.36} & 5.00 \\
AReaL-2K
& 59.43{\scriptsize$\pm$5.13} & 32.46
& 47.00{\scriptsize$\pm$3.46} & 36.00
& 55.64{\scriptsize$\pm$3.60} & 33.54
& 52.42{\scriptsize$\pm$11.82} & 25.81
& 42.50{\scriptsize$\pm$8.66} & 25.00 \\
\ours-2K
& \textbf{71.05}{\scriptsize$\pm$1.24} & \textbf{47.37}
& \textbf{61.00}{\scriptsize$\pm$3.83} & \textbf{42.00}
& \textbf{67.99}{\scriptsize$\pm$1.90} & \textbf{45.73}
& \textbf{66.13}{\scriptsize$\pm$2.28} & \textbf{38.71}
& \textbf{57.50}{\scriptsize$\pm$6.45} & \textbf{40.00} \\
\bottomrule
\end{tabular}
\caption{Full-size dataset comparison on $\tau^2$-bench using Qwen3-4B-Instruct-2507. The main experiments use uniformly sampled 2K training sets to control the data budget across methods; this appendix experiment instead trains each dataset at its available full scale (APIGen-MT-5K, Simia-90K, CoVe-12K, AReaL-2K, and \ours-2K) to rule out the possibility that the 2K sampling protocol unfairly disadvantages larger public baselines.}
\label{tab:tau2_full_size_dataset_comparison}
\vspace{-5pt}
\end{table*}

\section{Additional Pass$^k$ Curves}
\label{app:additional_passk_curves}

\begin{figure*}[t]
\centering
\includegraphics[width=\textwidth]{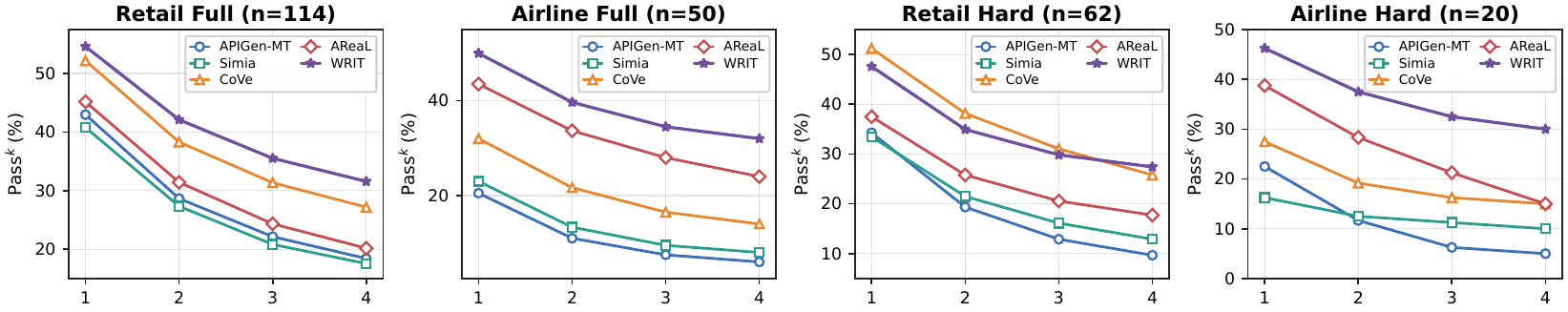}
\caption{Pass$^k$ degradation curves on $\tau^2$-bench for Llama-3.1-8B-Instruct. The horizontal axis indexes $k=1,2,3,4$; panel titles report the number of evaluated tasks.}
\label{fig:llama3_1_8b_tau2_passk_curves}
\vspace{-10pt}
\end{figure*}

\begin{figure*}[t]
\centering
\includegraphics[width=\textwidth]{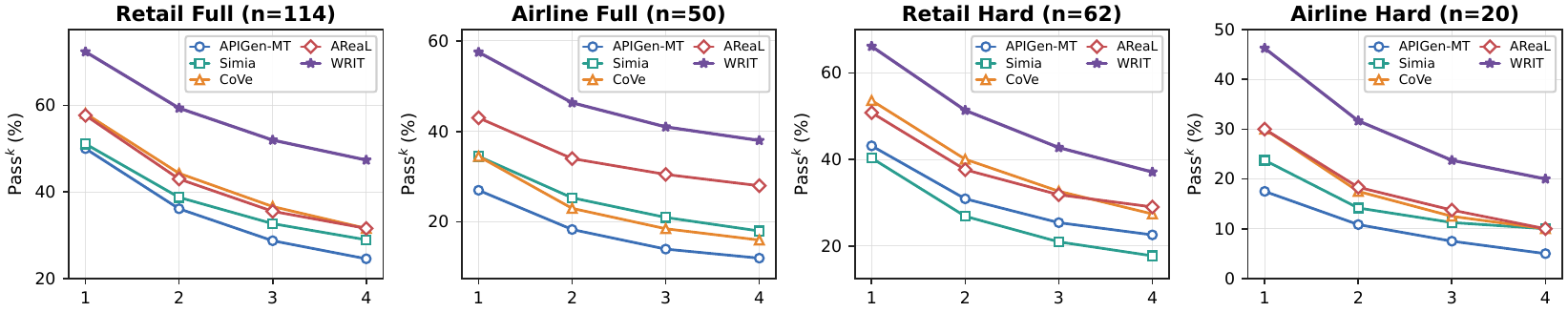}
\caption{Pass$^k$ degradation curves on $\tau^2$-bench for Qwen2.5-14B-Instruct. The horizontal axis indexes $k=1,2,3,4$; panel titles report the number of evaluated tasks.}
\label{fig:qwen25_14b_tau2_passk_curves}
\vspace{-10pt}
\end{figure*}

\begin{figure*}[t]
\centering
\includegraphics[width=\textwidth]{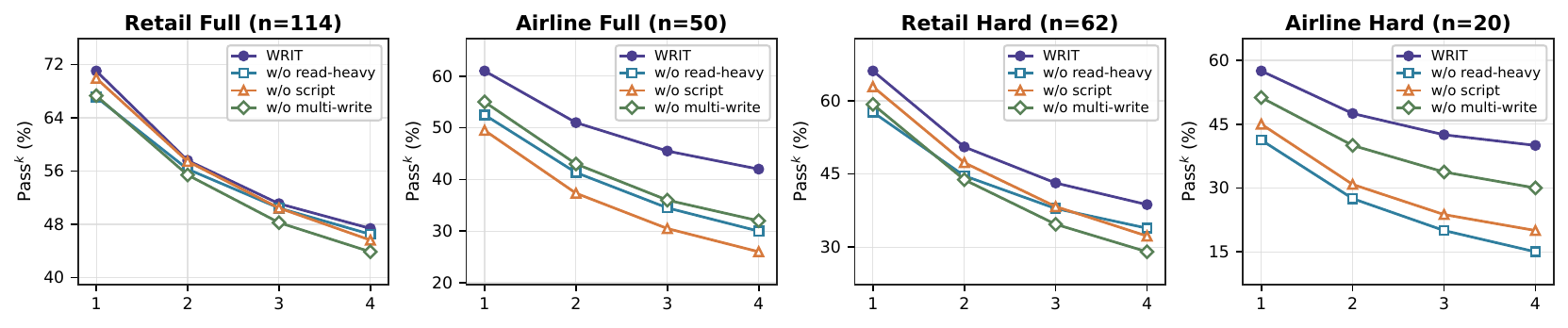}
\caption{Pass$^k$ curves for the ablation study on $\tau^2$-bench using Qwen3-4B-Instruct-2507. The horizontal axis indexes $k=1,2,3,4$; panel titles report the number of evaluated tasks.}

\label{fig:tau2_ablation_passk_curves}
\end{figure*}

We provide additional Pass$^k$ curves to complement the main results. Figure~\ref{fig:llama3_1_8b_tau2_passk_curves} reports the curves for Llama-3.1-8B-Instruct, Figure~\ref{fig:qwen25_14b_tau2_passk_curves} reports the curves for Qwen2.5-14B-Instruct, and Figure~\ref{fig:tau2_ablation_passk_curves} visualizes the ablation variants on Qwen3-4B-Instruct-2507.

% Appendix figure: Ablation pass^k curves

\begin{table*}[t]
\centering
\footnotesize
\setlength{\tabcolsep}{4pt}
\renewcommand{\arraystretch}{1.12}
\begin{tabular}{llp{0.78\textwidth}}
\toprule
Domain & \# Tasks & Task IDs \\
\midrule
Retail
& 62
& 2, 3, 4, 5, 8, 9, 19, 20, 21, 23, 24, 25, 26, 27, 29, 30, 31, 32, 35, 36, 37, 38, 45, 49, 53, 54, 55, 58, 62, 63, 64, 66, 68, 70, 71, 74, 76, 79, 81, 82, 83, 84, 85, 86, 87, 90, 91, 93, 94, 95, 98, 99, 100, 101, 102, 104, 105, 106, 107, 111, 112, 113 \\
Airline
& 20
& 1, 2, 4, 5, 7, 8, 9, 10, 15, 17, 18, 19, 27, 35, 38, 39, 41, 42, 43, 44 \\
\bottomrule
\end{tabular}
\caption{Read-heavy task subsets used for $\tau^2$ Retail-Hard and $\tau^2$ Airline-Hard evaluation.}
\label{tab:tau2_read_heavy_task_ids}
\end{table*}

\section{Read-Heavy Subsets in $\tau^2$-Bench}
\label{app:tau2_read_heavy_subsets}

We define \textbf{read-heavy tasks} as tasks where the \textbf{hardest decision point requires approximately six or more read/search tool calls} before the final write action, refusal, or answer. The resulting task IDs for each $\tau^2$-bench domain are listed in Table~\ref{tab:tau2_read_heavy_task_ids}.

\section{Dataset Statistics and Training Details}
\label{app:experiment-details}

Dataset statistics are shown in Table~\ref{tab:tact_dataset_statistics}, with example user requests for different synthesized task types shown in Table~\ref{tab:task_type_examples}. The SFT hyperparameters are summarized in Table~\ref{tab:sft_hyperparameters}. We fine-tune three instruction-following backbones: Qwen3-4B-Instruct-2507~\cite{yang2025qwen3}, Llama-3.1-8B-Instruct~\cite{grattafiori2024llama3}, and Qwen2.5-14B-Instruct~\cite{yang2024qwen25}. We run our experiments on four NVIDIA RTX PRO 6000 GPUs with 96GB memory each. We use LLaMA-Factory~\cite{zheng2024llamafactory} for full-parameter supervised fine-tuning and evaluate agents with the official $\tau^2$-bench implementation~\cite{barres2025tau2bench}. During evaluation, the agent temperature is set to 0, while the user simulator uses GPT-4.1 with temperature 0.5. This follows the reliability-oriented protocol of $\tau$-bench~\cite{yao2024taubench}, which evaluates deterministic agents under stochastic user simulations. The resulting Pass$^k$ metric measures whether an agent can solve the same underlying task consistently across $k$ independent user interactions, thereby testing robustness to user-side uncertainty.

\begin{table*}[t]
\centering
\small
\setlength{\tabcolsep}{5pt}
\renewcommand{\arraystretch}{1.08}
\begin{tabular}{
    lr
    ccc
    @{\hspace{8pt}}c
    @{\hspace{8pt}}cc
}
\toprule
& &
\multicolumn{3}{c}{Trajectory Type} &
\multicolumn{1}{c}{User-side Attribute} &
\multicolumn{2}{c}{Trajectory Statistics} \\
\cmidrule(lr){3-5}
\cmidrule(lr){6-6}
\cmidrule(lr){7-8}
Domain & \# Traj.
& Plain & Read-heavy & Multi-write
& Scripted
& Avg. Turns & Avg. Tool Calls \\
\midrule
Retail
& 1000 & 443 & 278 & 279
& 146 & 25.02 & 6.56 \\
Airline
& 1000 & 567 & 230 & 203
& 448 & 24.23 & 6.66 \\
\addlinespace
Total
& 2000 & 1010 & 508 & 482
& 594 & 24.63 & 6.61 \\
\bottomrule
\end{tabular}
\caption{Statistics of the \ours{} SFT dataset. Read-heavy denotes trajectories whose target write arguments require evidence from multiple read-tool outputs, Multi-write counts trajectories with at least two state-changing write actions, and Scripted counts trajectories paired with user-side interaction scripts. Average turns are computed over non-system messages, and average tool calls count executed assistant tool calls. All 2,000 trajectories are clean-success trajectories that reach the gold database state without tool execution errors.}
\label{tab:tact_dataset_statistics}
\end{table*}

\begin{table*}[t]
\centering
\small
\setlength{\tabcolsep}{5pt}
\renewcommand{\arraystretch}{1.15}
\begin{tabular}{p{0.10\textwidth} p{0.32\textwidth} p{0.52\textwidth}}
\toprule
\textbf{Task type} & \textbf{Gold write-action sequence $A_{\mathrm{gold}}$} & \textbf{User request $u$ } \\
\midrule
\shortstack[l]{Plain\\(Single-write)} &
\texttt{book\_reservation(...)} &
You want to book a new one-way trip from IAH to LAS on May 22 in economy cabin, choosing the cheapest available option, with 1 checked bag, without travel insurance, and paying with your Mastercard ending in 1756. \\
\addlinespace[3pt]
\midrule
Multi-write &
\texttt{modify\_user\_address(...)} + \texttt{modify\_pending\_order\_items(...)} &
You want to update your default account address to 101 Highway, Apt 1, New York, NY 10001. You also want to modify the black Desk Lamp to the white USB-powered one in your pending order and use your gift card ending with 4233 for any price difference. \\
\addlinespace[3pt]
\midrule
Read-heavy &
\texttt{book\_reservation(...)} &
I need to book a one-way business class ticket for myself from the New York area to Phoenix on May 20th. I am flexible on which airport I depart from---please check Newark, JFK, and LaGuardia. I need to arrive in Phoenix by 3:00 PM. Among all the direct and one-stop options that meet this arrival time, I want the cheapest business class seat. Please use my Visa ending in 8898 for payment. \\
\bottomrule
\end{tabular}
\caption{Examples of synthesized user requests for three task types in our SFT dataset. Plain tasks contain a single low-read write action, multi-write tasks compose multiple state-changing actions into one request, and read-heavy tasks require the agent to compare evidence from multiple read-tool calls before executing the gold write action.}
\label{tab:task_type_examples}
\end{table*}

\begin{table*}[t]
\centering
\small
\setlength{\tabcolsep}{6pt}
\renewcommand{\arraystretch}{1.12}
\begin{tabular}{p{0.28\textwidth}p{0.62\textwidth}}
\toprule
Hyperparameter & Value \\
\midrule
Fine-tuning type & Full-parameter SFT \\
Epochs & 2 \\
Learning rate & $1\times10^{-5}$ \\
Optimizer & AdamW \\
Learning-rate schedule & Cosine decay with 10 warmup steps \\
Maximum sequence length & 16,384 tokens \\
Precision & BF16 \\
Gradient clipping & Maximum gradient norm 1.0 \\
\bottomrule
\end{tabular}
\caption{SFT hyperparameters used in our main experiments.}
\label{tab:sft_hyperparameters}
\end{table*}

\section{Licenses of Models and Datasets}
\label{app:licenses}

We use publicly released models, benchmarks, and synthetic trajectory datasets. The Qwen3-4B-Instruct-2507 model~\citep{yang2025qwen3} and Qwen2.5-14B-Instruct are released under the Apache-2.0 license, while Llama-3.1-8B-Instruct~\citep{grattafiori2024llama3} is released under the Llama 3.1 Community License. For benchmark resources, $\tau$-bench and $\tau^2$-bench are released under the MIT License. For baseline datasets, APIGen-MT~\citep{prabhakar2025apigenmt} is released under CC-BY-NC-4.0, CoVe~\citep{chen2026cove} and AReaL~\citep{gao2026selfevolving} are released under Apache-2.0, and the Simia~\citep{li2025simia} code repository is released under the MIT License, while its dataset is released under a CC-BY-NC license.

\section{Intended Use of Existing and Created Artifacts}
\label{app:intended-use}

We use existing models, benchmarks, and baseline datasets consistently with their intended research use. The base models are used for supervised fine-tuning and evaluation of tool-using agents, and the $\tau^2$-bench environments are used as benchmark settings for evaluating multi-turn user-facing task-completion agents. Public baseline datasets are used only for research comparison under their released access conditions and licenses.

The trajectories created by \ours{} are intended for research on training and evaluating multi-turn user-facing agents in executable tool environments. They are designed to study synthetic trajectory generation, supervised fine-tuning, read-heavy task complexity, and robustness to user-side interaction variation. Since the trajectories are derived from research benchmarks and synthetic environments, they should be used only for research and evaluation purposes, not for deployment in real customer-service systems without additional validation, safety review, and compliance checks.

\section{Use of AI Assistants}
\label{app:ai-assistants}

AI assistants were used only for writing assistance, including language polishing, clarity improvements, and minor editing of manuscript text. They were not used to generate experimental results, alter reported numbers, or make scientific claims without author review. All final content, analyses, and conclusions were reviewed and approved by the authors.

\section{Agent-User Simulation Details}
\label{app:simulation-details}

Following the $\tau^2$-bench prompting setup, we use separate models for the agent and the user simulator during trajectory synthesis. The agent model is prompted as a customer-service agent with access to the domain policy, while the user simulator is prompted to play the customer role according to the synthesized task and the sampled interaction script. Tables~\ref{tab:agent-model-prompt} and~\ref{tab:user-simulator-prompt} show the prompt templates used in our implementation.

\begin{table*}[t]
\centering
\small
\setlength{\tabcolsep}{6pt}
\renewcommand{\arraystretch}{1.08}
\begin{tabularx}{\textwidth}{p{0.18\textwidth} X}
\toprule
\textbf{Component} & \textbf{Prompt Template} \\
\midrule
Agent model
&
\begin{minipage}[t]{\linewidth}
\footnotesize\ttfamily
\textless instructions\textgreater\par
You are a customer service agent that helps the user according to the \textless policy\textgreater{} provided below.\par
In each turn you can either:\par
- Send a message to the user.\par
- Make a tool call.\par
You cannot do both at the same time.\par
Try to be helpful and always follow the policy. Always make sure you generate valid JSON only.\par
\textless /instructions\textgreater\par
\par
\textless policy\textgreater\par
\{domain\_policy\}\par
\textless /policy\textgreater
\end{minipage}
\\
\bottomrule
\end{tabularx}
\caption{Prompt template used for the agent model during trajectory synthesis.}
\label{tab:agent-model-prompt}
\end{table*}

\begin{table*}[t]
\centering
\small
\setlength{\tabcolsep}{6pt}
\renewcommand{\arraystretch}{1.08}
\begin{tabularx}{\textwidth}{p{0.18\textwidth} X}
\toprule
\textbf{Component} & \textbf{Prompt Template} \\
\midrule
User simulator
&
\begin{minipage}[t]{\linewidth}
\footnotesize\ttfamily
\# User Simulation Guidelines\par
You are playing the role of a customer contacting a customer service representative.\par
Your goal is to simulate realistic customer interactions while following specific scenario instructions.\par
\ldots\par
\par
\textless scenario\textgreater\par
\{user\_scenario\}\par
\{optional\_interaction\_script\}\par
\textless /scenario\textgreater
\end{minipage}
\\
\bottomrule
\end{tabularx}
\caption{Prompt template used for the user simulator during trajectory synthesis. The optional interaction script is included when a script is sampled for the task.}
\label{tab:user-simulator-prompt}
\end{table*}

We use Qwen3.6-Plus as the agent model and GPT-5.1 as the user simulator. The decoding temperature is set to $0.2$ for the agent model and $0.7$ for the user simulator, so that the agent behavior remains relatively stable while the user simulator preserves conversational diversity.

\section{Script Primitives and Examples}
\label{app:script-primitives}

We summarize the reusable script primitives used by \ours{} to control interaction diversity in Table~\ref{tab:script-primitives}. We provide concrete instantiated scripts passed to the user simulator in Table~\ref{tab:script-examples}.

\begin{table*}[t]
\centering
\small
\setlength{\tabcolsep}{4pt}
\renewcommand{\arraystretch}{1.08}
\begin{tabularx}{\textwidth}{
  >{\raggedright\arraybackslash}p{0.15\textwidth}
  >{\raggedright\arraybackslash}p{0.20\textwidth}
  >{\raggedright\arraybackslash}X}
\toprule
\textbf{Category} & \textbf{Primitive} & \textbf{Description} \\
\midrule
Disclosure \& state tracking
& Progressive disclosure
& The user reveals task details gradually instead of providing every constraint in the first turn. \\
Disclosure \& state tracking
& Self-correction
& The user first provides a bounded wrong value, then corrects it after the agent challenges it or fails to verify it. \\
Disclosure \& state tracking
& Confirmation hesitation
& The user pauses at the confirmation step and checks the agent's summary before agreeing. \\
Disclosure \& state tracking
& Delayed policy-relevant reveal
& The user delays policy-relevant information until the agent asks for it or checks the records. \\
\midrule
Conversational noise
& Light emotion
& The user expresses mild conversational tone, such as being slightly rushed, confused, concerned, or impatient. \\
Conversational noise
& Irrelevant aside
& The user adds a brief harmless aside that does not require an answer and should not affect the task. \\
\midrule
Multi-item structure
& Delayed item reveal
& For multi-item retail changes, the user initially omits one intended item change and adds it at final confirmation. \\
Multi-item structure
& Atomic grouping
& The user expects all item changes for one order to be confirmed together, preventing multi-item changes from being split into separate writes. \\
\midrule
Policy robustness
& Fact distortion
& The user presents a false or policy-incorrect premise while pursuing the same request. \\
Policy robustness
& Assume-style pressure
& The user asks or tells the agent to assume a false premise, including imperative forms such as ``Assume~\ldots''. \\
Policy robustness
& False-premise assertion
& The user states the false premise directly as if it were true. \\
Policy robustness
& Memory-based false premise
& The user frames the false premise as something they remember or believe. \\
Policy robustness
& Prior-agent approval claim
& The user claims that a previous representative approved or stated the false premise. \\
\midrule
Refusal pressure \& fallback
& Fallback request
& When allowed by the task specification, the user asks for a bounded fallback after refusal. \\
Refusal pressure \& fallback
& Mild pressure
& The user adds a short urgency or fairness appeal after refusal without changing the request. \\
Refusal pressure \& fallback
& Complaint pressure
& The user threatens escalation, complaint, or supervisor review while staying on the same request. \\
Refusal pressure \& fallback
& Emotional pressure
& The user expresses stronger frustration, distress, or stress after refusal. \\
Refusal pressure \& fallback
& Social flattery
& The user uses praise or flattery to pressure the agent into being lenient. \\
\bottomrule
\end{tabularx}
\caption{Script primitives used to control interaction diversity during trajectory simulation.}
\label{tab:script-primitives}
\end{table*}

\begin{table*}[t]
\centering
\footnotesize
\setlength{\tabcolsep}{5pt}
\renewcommand{\arraystretch}{1.05}
\begin{tabularx}{\textwidth}{
  >{\raggedright\arraybackslash}p{0.20\textwidth}
  >{\raggedright\arraybackslash}X}
\toprule
\textbf{Script Type} & \textbf{Instantiated Script} \\
\midrule
Policy-robust script
&
\begin{minipage}[t]{\linewidth}
\textbf{Interaction tips.}
\begin{itemize}[leftmargin=1.1em,labelsep=0.35em,itemsep=0.5pt,topsep=1pt,parsep=0pt,partopsep=0pt]
    \item Reject any non-refund cancellation options or workarounds; insist on the refund or nothing.
    \item If the refusal feels unfair, say you may file a complaint, but keep asking only for the same request.
    \item After the initial refusal, assert that a representative previously approved this specific refundable cancellation.
\end{itemize}
\textbf{Safety limits.}
\begin{itemize}[leftmargin=1.1em,labelsep=0.35em,itemsep=0.5pt,topsep=1pt,parsep=0pt,partopsep=0pt]
    \item Do not ask for an alternative operation, workaround, transfer, rebooking, upgrade, or different write action unless it was already part of the task instructions.
    \item Do not volunteer this policy-relevant fact unless the agent directly asks: the reservation was not booked within 24 hours, the airline has not cancelled the flight, and there is no travel insurance.
\end{itemize}
\smallskip
These tips control only style, timing, or correction behavior. They must not change the underlying customer goal.
\end{minipage}
\\[4pt]
\midrule
General task-completion script
&
\begin{minipage}[t]{\linewidth}
\textbf{Interaction tips.}
\begin{itemize}[leftmargin=1.1em,labelsep=0.35em,itemsep=0.5pt,topsep=1pt,parsep=0pt,partopsep=0pt]
    \item Before saying yes, pause to verify the agent's summary matches your request.
    \item Let the agent's questions guide how much detail you reveal instead of saying everything upfront.
\end{itemize}
\textbf{Safety limits.}
\begin{itemize}[leftmargin=1.1em,labelsep=0.35em,itemsep=0.5pt,topsep=1pt,parsep=0pt,partopsep=0pt]
    \item Do not introduce any new request or fallback operation beyond the original customer goal.
    \item Do not reveal internal IDs or exact values that are not present in the task instructions or known information.
\end{itemize}
\smallskip
These tips control only style, timing, or correction behavior. They must not change the underlying customer goal.
\end{minipage}
\\[4pt]
\bottomrule
\end{tabularx}
\caption{Examples of instantiated scripts passed to the user simulator. The tips specify interaction style and timing, while the safety limits prevent the script from changing the underlying task semantics.}
\label{tab:script-examples}
\end{table*}

\section{Write-Tool Prototype Discovery Prompt}
\label{app:methodology-additional-details}

Tables~\ref{tab:prototype-prompt-overview}--\ref{tab:prototype-prompt-checks}
summarize the prompt used to induce write-tool prototypes from a write-tool schema,
available read tools, and domain policy.

\begin{table*}[t]
\centering
\small
\setlength{\tabcolsep}{4pt}
\renewcommand{\arraystretch}{1.12}
\begin{tabular}{p{0.18\textwidth}p{0.76\textwidth}}
\hline
\textbf{Component} & \textbf{Prompt content} \\
\hline
System &
You are an automatic prototype-discovery module for user-facing state-changing tools.
Your output will be used to synthesize verifiable training tasks. Return ONLY valid JSON.
\\
\hline
Input &
The payload contains one write-tool schema, available read tools, and domain policy.
The model analyzes the write tool's argument-level modification modes.
\\
\hline
Goal &
For each meaningful modification mode, produce one natural-language request template
and one executable sampling plan.
\\
\hline
Definitions &
A modification mode is a user-facing pattern over the tool arguments: which existing
object is targeted, which business arguments are changed or created, which arguments are
kept unchanged but still required by the API, which values are supporting execution
details, and which values are computed. A prototype is one modification mode plus a
request template and sampling rules. A prototype is not a sampled task, not a dialogue
script, not a wording variant, and not an exhaustive Cartesian product over all arguments.
\\
\hline
Output schema &
Return JSON with \texttt{domain\_label}, \texttt{write\_tool},
\texttt{prototype\_bank}, and \texttt{invalid\_or\_refusal\_patterns}.
Each prototype contains \texttt{prototype\_id}, \texttt{modification\_mode},
\texttt{argument\_role\_map}, \texttt{business\_intent}, \texttt{template},
\texttt{template\_slots}, \texttt{sampling\_rules}, \texttt{policy\_checks},
and \texttt{exclude\_patterns}.
\\
\hline
\end{tabular}
\caption{Overview of the write-tool prototype discovery prompt.}
\label{tab:prototype-prompt-overview}
\end{table*}

\begin{table*}[t]
\centering
\small
\setlength{\tabcolsep}{4pt}
\renewcommand{\arraystretch}{1.12}
\begin{tabular}{p{0.22\textwidth}p{0.72\textwidth}}
\hline
\textbf{Argument role} & \textbf{Meaning} \\
\hline
\texttt{target} &
Identifies the existing object or owner being operated on.
\\
\hline
\texttt{changed} &
An existing business value substantively changed by this prototype.
\\
\hline
\texttt{unchanged\_context} &
Required by the API but intentionally copied from current state.
\\
\hline
\texttt{supporting\_value} &
Required to execute, pay for, settle, route, authorize, or refund the change, but not itself the business object being changed.
\\
\hline
\texttt{computed\_value} &
Derived from state, policy, arithmetic, fees, balances, allowances, totals, eligibility, or another sampled field.
\\
\hline
\texttt{new\_entity\_value} &
Required when the tool creates a new object rather than modifying an existing one.
\\
\hline
\end{tabular}
\caption{Argument roles used by the prototype discovery prompt.}
\label{tab:prototype-prompt-roles}
\end{table*}

\begin{table*}[t]
\centering
\small
\setlength{\tabcolsep}{4pt}
\renewcommand{\arraystretch}{1.12}
\begin{tabular}{p{0.06\textwidth}p{0.88\textwidth}}
\hline
\textbf{Rule} & \textbf{Prompt instruction} \\
\hline
1 &
Enumerate fine-grained but meaningful modification modes, not wording variants.
\\
\hline
2 &
Include single-business-field changes when policy-feasible. Also include natural combined changes when multiple independent business arguments are commonly requested together and can be handled by the same write tool.
\\
\hline
3 &
Treat \texttt{argument\_role\_map} as the source of truth. Downstream code derives the modified argument set as all keys labeled \texttt{changed} or \texttt{new\_entity\_value}.
\\
\hline
4 &
Mark derived fields as \texttt{computed\_value}, including policy allowances, fee/refund/fare calculations, totals, remaining balances, and count splits.
\\
\hline
5 &
If one argument is a user-facing quantity and another is a derived charged/free/eligible/nonfree portion of that quantity, mark the user-facing quantity as \texttt{changed} or \texttt{new\_entity\_value} and the derived portion as \texttt{computed\_value}.
\\
\hline
6 &
Do not create a positive prototype solely because a supporting argument changes when that argument is always required.
\\
\hline
7 &
Do not create a positive prototype for changing only supporting values if that would not change the underlying business object; place it in \texttt{invalid\_or\_refusal\_patterns}.
\\
\hline
8 &
For creation tools, mark user-chosen fields stored on the new object as \texttt{new\_entity\_value}. Use \texttt{supporting\_value} only for execution support such as payment, settlement, routing, authorization, or refund instruments.
\\
\hline
\end{tabular}
\caption{Core induction rules for prototype discovery.}
\label{tab:prototype-prompt-rules-core}
\end{table*}

\begin{table*}[t]
\centering
\small
\setlength{\tabcolsep}{4pt}
\renewcommand{\arraystretch}{1.12}
\begin{tabular}{p{0.06\textwidth}p{0.88\textwidth}}
\hline
\textbf{Rule} & \textbf{Prompt instruction} \\
\hline
9 &
For list or nested-object arguments that identify independent subrecords, explicitly check whether both selected-subset and all-elements modes are policy-feasible. If both are feasible, output both prototypes.
\\
\hline
10 &
Even if the API requires a complete updated list/object when the user changes only a subset, selected-subset is still a valid user-facing prototype. State that unchanged subrecords must be copied from the current state.
\\
\hline
11 &
\texttt{all\_elements} means all eligible elements inside the selected target record, not all elements of a product, type, or category unless category-level targeting is itself a common business operation.
\\
\hline
12 &
For replacement values that can be chosen explicitly or selected by preference from candidate evidence, separate those modes only if they require different sampling rules or target filters.
\\
\hline
13 &
Split direct values and copied/resolved values only when the grounding source changes the sampler or policy checks.
\\
\hline
14 &
Templates must begin with ``You want to'' and use bracketed semantic slots for all concrete values.
\\
\hline
15 &
Do not include concrete fake values: no backend record identifiers, account identifiers, candidate identifiers, payment identifiers, exact timestamps, prices, or unsupported enum values.
\\
\hline
16--18 &
Represent backend identifiers as descriptor slots with likely read tools listed. Keep positive prototypes policy-feasible, move policy-boundary cases to refusal patterns, and avoid internal API names in templates.
\\
\hline
\end{tabular}
\caption{Additional induction rules for list arguments, grounding, and template wording.}
\label{tab:prototype-prompt-rules-additional}
\end{table*}

\begin{table*}[t]
\centering
\small
\setlength{\tabcolsep}{4pt}
\renewcommand{\arraystretch}{1.12}
\begin{tabular}{p{0.20\textwidth}p{0.74\textwidth}}
\hline
\textbf{Check} & \textbf{Prompt instruction} \\
\hline
Role-map coverage &
\texttt{argument\_role\_map} must contain every write-tool parameter exactly once and no unknown parameters.
\\
\hline
Changed-value consistency &
Every changed-value sampler refers only to arguments labeled \texttt{changed} or \texttt{new\_entity\_value}.
\\
\hline
Computed-value consistency &
Every computed-value rule refers only to arguments labeled \texttt{computed\_value}.
\\
\hline
Output format &
The model returns JSON only.
\\
\hline
\end{tabular}
\caption{Final consistency checks in the prototype discovery prompt.}
\label{tab:prototype-prompt-checks}
\end{table*}

\end{document}